\documentclass[review]{elsarticle}

\usepackage{lineno}
\modulolinenumbers[5]

\usepackage{algorithm}			% for algorithm
\usepackage{algpseudocode}	% for algorithmic
\usepackage{amsmath}			% for eqref, etc.
\usepackage{bm}				% for boldsymbol
\usepackage{gensymb}			% for degree
\usepackage{graphicx}			% for includegraphics and graphicspath
\usepackage{url}

\graphicspath{{figures/}{}}
\journal{Pattern Recognition}
\newcommand{\bbeta}{{\boldsymbol{\beta}}}
\newcommand{\bmu}{{\boldsymbol{\mu}}}
\newcommand{\bp}{{\boldsymbol{p}}}
\newcommand{\bx}{{\boldsymbol{x}}}

\begin{document}
\begin{frontmatter}

\title{Parameterized Principal Component Analysis}

\author[mymainaddress]{Ajay Gupta}
\author[mymainaddress]{Adrian Barbu\corref{mycorrespondingauthor}}
\address{}

\ead[url]{http://ani.stat.fsu.edu/~abarbu/}

\cortext[mycorrespondingauthor]{Corresponding author.}
\ead{abarbu@stat.fsu.edu}

\address[mymainaddress]{Department of Statistics, Florida State University, USA}

\begin{abstract}
\par{When modeling multivariate data, one might have an extra parameter of contextual information that could be used to treat some observations as more similar to others.
For example, images of faces can vary by age, and one would expect the face of a 40 year old to be more similar to the face of a 30 year old than to a baby face.}
\par{We introduce a novel manifold approximation method, parameterized principal component analysis (PPCA) that models data with linear subspaces that change continuously according to the extra parameter of contextual information (e.g. age), instead of ad-hoc atlases.
Special care has been taken in the loss function and the optimization method to encourage smoothly changing subspaces across the parameter values.
The approach ensures that each observation's projection will share information with observations that have similar parameter values, but not with observations that have large parameter differences.}
\par{We tested PPCA on artificial data based on known, smooth functions of an added parameter, as well as on three real datasets with different types of parameters.
We compared PPCA to PCA, sparse PCA and to independent principal component analysis (IPCA), which groups observations by their parameter values and projects each group using PCA with no sharing of information for different groups.
PPCA recovers the known functions with less error and projects the datasets' test set observations with consistently less reconstruction error than IPCA does. In some cases where the manifold is truly nonlinear, PCA outperforms all the other manifold approximation methods compared.}
\end{abstract}
\end{frontmatter}

\section{Introduction}
In recent years, storing and modeling multidimensional data have become very common.
Potential datasets include different attributes of potential customers, multiple currencies' exchange rates for each day, and vectorized images.
Although these data often lie on non-linear manifolds, a linear manifold or a combination of linear manifolds can often provide a practical and suitably accurate approximation.
Particularly for data of very high dimensionality such as vectorized images, a model may need to produce a reduced-dimension representation of the original observations. 

\noindent {\bf Generic manifold methods} only use the observations to approximate the manifold, without any extra information. Constructing the manifold from the data requires constructing a coordinate system on the manifold and projecting the observations onto the manifold. These tasks could be challenging if no side information is available.

 One popular and effective technique for modeling linear manifolds and incorporating dimensionality reduction is principal component analysis (PCA), which finds a basis $\boldsymbol{P}$ of vectors that can capture the highest-variance directions from the original data \cite{martinez2001pca}.

\par{Pitelis et al. (2013) showed how an ``atlas'' of overlapping linear manifolds that they labeled ``charts" could model a non-linear manifold very effectively \cite{pitelis2013learning}.
Their model was learned by a hill-climbing approach which alternated between assigning observations to charts based on the observations' values and refitting each chart using PCA performed on the relevant subset of observations.
The initial charts, which were necessary for the first assignments, could be found by PCA on bootstrap samples. The number of charts was selected by the method based on a user-supplied penalty $\lambda$.}
\par{Vidal, Ma, and Sastry (2005) introduced Generalized Principal Component Analysis (GPCA), which similarly addressed the idea of dividing a larger manifold into multiple local manifolds.
GPCA used polynomials based on Veronese maps to modify and combine elements of the original data vectors. GPCA could still learn the coefficients of the monomial terms by PCA, though, because the relationship between the full polynomial and these coefficients was still linear \cite{vidal2005generalized}.
The experimental success of GPCA showed that multiple applications of (linear) PCA could be used to learn a complicated manifold, although the local manifolds learned were typically non-linear.
The authors noted, though, that piecewise linear models (which could be learned by multiple PCA applications without GPCA's polynomials) are ``excellent" in many practical applications at balancing the need for model expressiveness with the desire for model simplicity \cite{vidal2003generalized}.
Like the atlas method, GPCA could also select the appropriate number of local manifolds, but using the ranks of Veronese maps evaluated on the observations instead of user-supplied parameters \cite{vidal2005generalized}.}

The Joint Parameterized Atom Selection (PATS) method \cite{vural2013learning} learns Pattern Transformation Manifolds (PTM), which are manifolds of images undergoing a family of geometric transformations. The PATS method therefore is designed for a different purpose than our method, since our work can be applied to manifolds that are not PTMs. In particular, the PATS method cannot be applied to the examples that will be presented in Section \ref{section:experiments}.

 Other related works include Sparse PCA \cite{zou2006sparse} where observations are represented as linear combinations of a small number of basis vectors and its precursor SCoTLASS \cite{jolliffe2003modified}, where the sparse PCA coefficients are obtained through $L_1$ constraints. In Joint Sparse PCA \cite{yi2017joint} the authors modify the Sparse PCA loss function to simultaneously obtain feature selection and Sparse PCA.
A generalization of PCA to tensor data was presented by 
Multilinear PCA \cite{lu2006multilinear}, while Multilinear Sparse PCA \cite{lai2014multilinear} generalizes Sparse PCA to tensor data. 

\noindent {\bf Manifold learning with side information.} Techniques such as GPCA and the atlas-based method exist for identifying local manifolds for observations
based on the observations' values, and for identifying the number of local manifolds.
Situations exist, however, in which data are thought to lie approximately on local manifolds that estimate a larger manifold, but one knows some extra information about which local manifold corresponds to each observation instead of needing the algorithm to discover this. This extra information could be discrete or continuous.
For example, one may be modeling images of vehicles using a known class (``car," ``motorcycle," ``SUV," or ``truck") for each observation, or modeling face images where the age of the person is also known.

\begin{figure}[t]
\centering
\includegraphics[width=12cm]{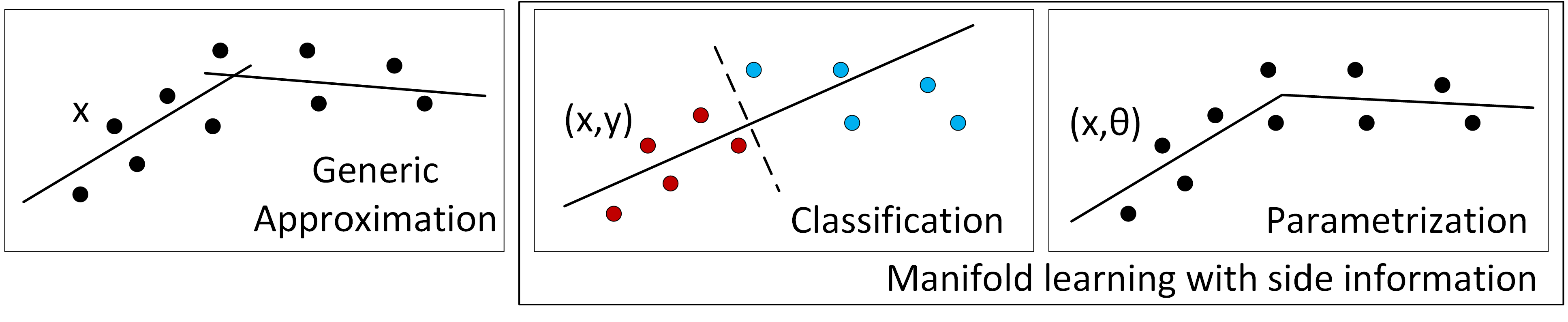}
\vskip -4mm
\caption{Illustration of different manifold learning methods and their goals.}
\label{fig:manifolds}
\vspace{-4mm}
\end{figure}\noindent {\bf Classification: learning manifolds using class information.} When the side information is in the form of discrete class labels, it is often desired to model the manifolds for each class for the best separation between the classes. While the generic methods discussed above had the goal of manifold approximation, these methods have a different goal: classification, as illustrated in Figure \ref{fig:manifolds}. 

In this respect, linear discriminant analysis (LDA) uses linear manifolds to perform dimensionality reduction to best separate the classes rather than to capture the variation of each class \cite{martinez2001pca}. Modified PCA \cite{luo2003modified} improves face recognition by dividing the PC coefficients to the square root of their corresponding eigenvalues. 
Other techniques such as Class-Information-Incorporated Principal Component Analysis \cite{chen2005class},  Locality Preserving Projections \cite{niyogi2004locality}, Multi-Manifold Semi-Supervised Learning \cite{goldberg2009multi}, Multimodal Oriented Discriminant Analysis \cite{de2005multimodal}, and Semi-Supervised Dimensionality Reduction \cite{zhang2007semi} learn manifolds in the presence of classes.
Like LDA, they are focused on classification to a local manifold rather than focused on modeling the observations once the classes are known.

These two goals of manifold learning - approximation vs classification - are clearly mentioned in the Joint Parameterized Atom Selection (PATS) paper \cite{vural2013learning} as two different objectives for building a manifold.

\noindent {\bf Parameterization: learning manifolds with a continuous context parameter.} In this work we are interested in modeling manifolds when the side information is continuous in the form of a context parameter $\theta$. This side information can help more accurately project each observation to the correct local manifold, obtaining a more accurate manifold approximation.

\par{Various applications exist in which this extra contextual information would be continuous.
For vehicle images, one could model them differently based on the vehicles' weights, volumes, or prices (MSRPs).
For face images one could model them differently based on their age, 3D pose, and illumination.
Daily percent changes in a stock's closing share price could use the stock's market capitalization,
because smaller-capitalization stocks are thought to have more volatile price movements.
Lenders with multiple recorded attributes about their borrowers could use the borrowers' rates of interest or
FICO credit scores as the side information.}

\par{One reason that modeling with continuous side information has been effectively unaddressed is that one could discretize the side information into a number of groups and treat the modeling problem as many separate problems, each of which could be addressed by existing techniques such
as PCA.
For notational simplicity, we will refer to the use of a separate PCA model for each group as Independent
Principal Component Analysis (IPCA), because none of the groups' models use information from the observations of the other
groups.}

\par{This contextual parameter carries ordinal and interval information that would be ignored by IPCA.
Consider borrower data such as a borrower's number of late payments, with credit scores as the parameter,
and bins 300-350, 350-400, and so on until 800-850.
The average late payments might decrease in the training examples as the bin increases,
except that the 400-450 bin might have a surprisingly low average.
IPCA would ignore the other bins and the pattern they form,
which could overfit the training examples in the 400-450 bin.
Additionally, IPCA would treat customers with credit scores of 355 and 845 as completely different from one
with a credit score of 345, because all three are in different bins.
It would ignore that the difference between 345 and 355 is much smaller than the difference between 345 and
845, rather than enforcing similarity between how the 345-score and 355-score observations are modeled.}
\par{In this paper, we propose a new method called parameterized principal component analysis (PPCA) for
creating a PCA-like linear model for multivariate data that is continuously changing with a separate parameter with
known, observation-specific values.

Like IPCA, PPCA makes multiple linear models of mean vectors and bases, which are based on known divisions of the parameter space.
Unlike IPCA, PPCA interpolates between the points in the parameter space at which mean vectors and bases
were fitted, and penalizes differences in the models for similar parameter values. For example the PPCA manifold between two consecutive bin endpoints in 3D with a single principal vector would be a ruled surface, such as shown in Figure \ref{fig:reg}, right.

\begin{figure}[htb]
\centering
\includegraphics[height=4.cm]{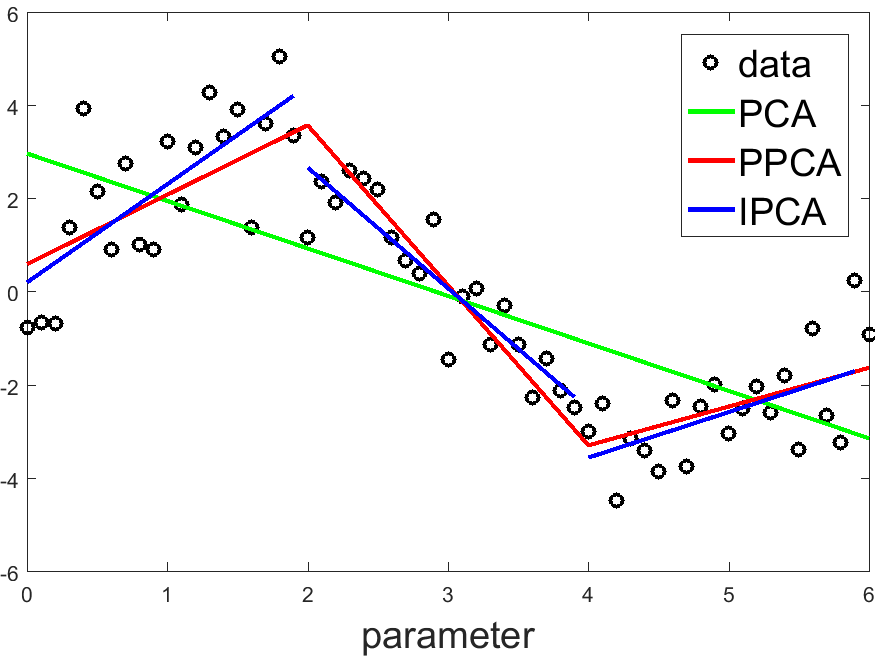}
\hspace{3mm}
\includegraphics[height=4.cm]{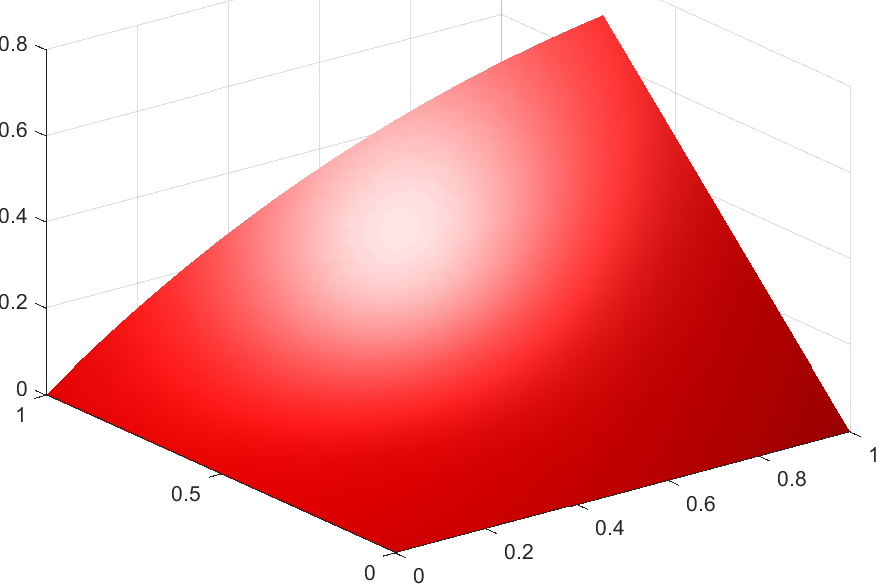}
\vskip -4mm
\caption{Left: Illustration of the difference between PCA, PPCA and IPCA. Right: a ruled surface is obtained by linear interpolation of the corresponding principal vectors at the bin endpoints.}
\label{fig:reg}
\vspace{-3mm}
\end{figure}
In Figure \ref{fig:reg}, left, are illustrated the differences between PCA, IPCA and PPCA on a simple 1D data and three bins for the parameter values. In PCA, a simple linear model is fit through all the data. In IPCA, separate models are fit independently on the data from each bin. In PPCA, three linear models are fit but enforced to match at bin endpoints. This type of continuity is enforced for both the mean vectors and for the principal directions.

We describe the PPCA model in Section \ref{section:ppcageneral}, and discuss its implementation
in Sections \ref{section:ppcafitting} and \ref{section:ppcageneralizations}.
In Section \ref{section:experiments}, we apply PPCA to artificial data following smooth functions of the
parameter, and to three real datasets: shapes of differently-sized lymph nodes, human facial images with
different degrees of added blurriness, and human facial images with different angles of yaw rotation.
In all four experiments, PPCA outperformed IPCA, and was particularly beneficial when the number of training
examples was limited.}

\section{Parameterized Principal Component Analysis}
\label{section:ppcageneral}
\begin{figure}[htb]
\centering
\includegraphics[height=5.cm]{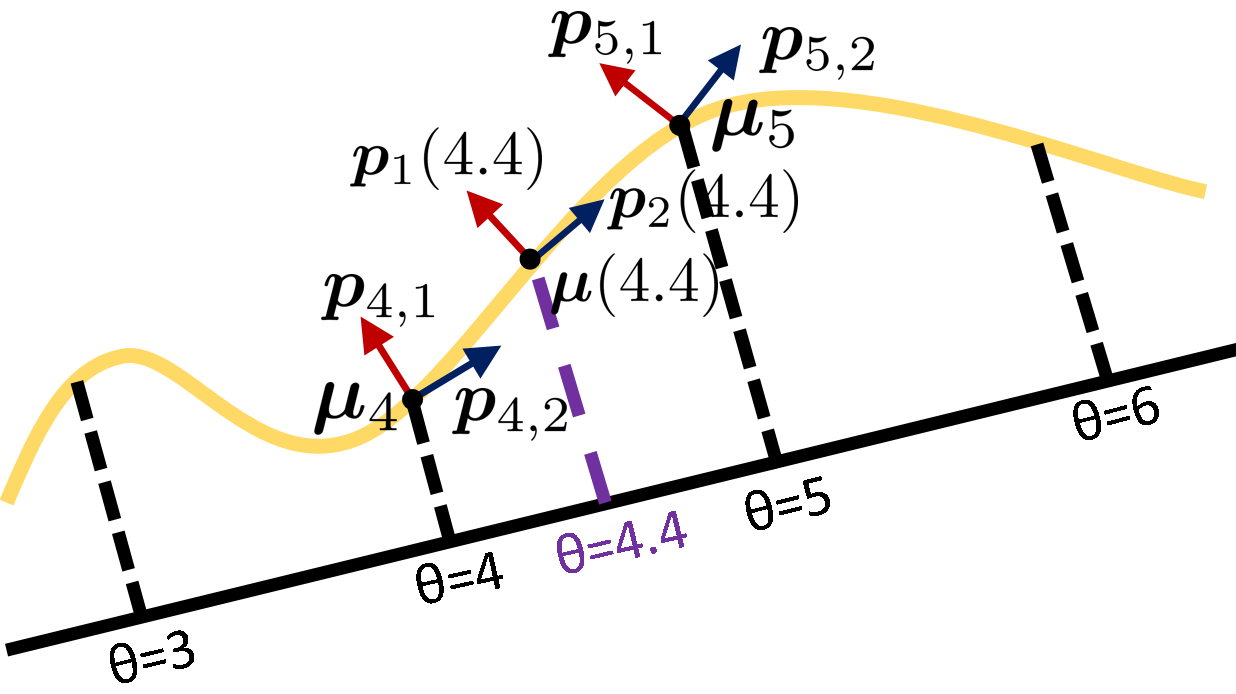}
\vskip -6mm
\caption{Example of a manifold represented by PPCA together with the bin means $\bmu_p$ and principal vectors $\bp_{b,i}$.}
\label{fig:manifold}
\vspace{-3mm}
\end{figure}
\par{Parameterized principal component analysis (PPCA) applies to an environment in which there are $n$ observations $\boldsymbol{x}_i$, each of dimension $K$, and there are $B$ bin endpoints arising from the $B-1$ bins that partition the acceptable range of the parameter $\theta$.
Each bin endpoint $b$ has a mean vector $\boldsymbol{\mu}_b$ and $V$ basis vectors $\boldsymbol{p}_{b,v}$.
Each bin endpoint corresponds to a value of $\theta$, and an observation $\boldsymbol{x}_i$'s parameter value $\theta_i$ dictates $\boldsymbol{x}_i$'s bin, with lower endpoint $b_{(l),i}$ and upper endpoint $b_{(u),i}$.
Figure \ref{fig:manifold} shows an example using a parameter that varies from $\theta=3$ to $\theta=6$.
Note that a bin endpoint usually applies to two bins.
For the example in Figure \ref{fig:manifold}, the bin endpoint at $\theta=5$ would be an endpoint for the 4-5 bin and for the 5-6 bin.}
\par
\begin{figure}
\centering
\includegraphics[scale=0.5]{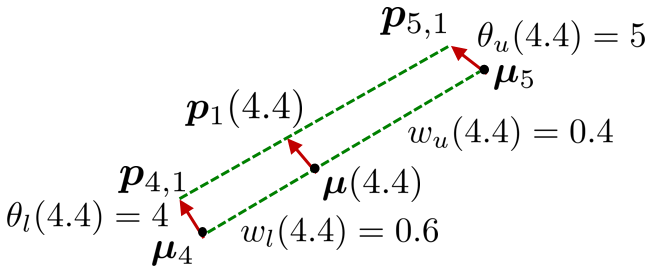}
\vskip -4mm
\caption{Example of combination from bin endpoint weights}
\label{fig:weights}
\vspace{-3mm}
\end{figure}

{The parameter $\theta_i$ can be translated into weights $w_l(\theta_i)$ and $w_u(\theta_i)$ for bin endpoints $b_{(l),i}$ and $b_{(u),i}$.
Equation \eqref{eq:weights} shows this, using $\theta_l(\theta_i)$ and $\theta_u(\theta_i)$ as the parameter values for the bin's lower and upper endpoint, respectively.
Figure \ref{fig:weights} shows an example for an observation with $\theta_i=4.4$.
It has a 60\% weight for the bin endpoint at $\theta=4$ and a 40\% weight for the bin endpoint at $\theta=5$, because 4.4 is 60\% of the way from 5 to 4, and 40\% of the way from 4 to 5.
\begin{equation}
\label{eq:weights}
w_l(\theta_i) = \frac{\theta_u(\theta_i)-\theta_i}{\theta_u(\theta_i)-\theta_l(\theta_i)},
w_u(\theta_i) = \frac{\theta_i-\theta_l(\theta_i)}{\theta_u(\theta_i)-\theta_l(\theta_i)}
\end{equation}
These weights can produce a mean vector $\boldsymbol{\mu}(\theta_i)$ and a basis $\boldsymbol{P}(\theta_i)$ that are specific to the observation's parameter $\theta_i$, as shown in Equations \eqref{eq:weightedmean} and \eqref{eq:weightedvectors}.
\begin{equation}
\label{eq:weightedmean}
\boldsymbol{\mu}(\theta_i) = w_l(\theta_i) \boldsymbol{\mu}_{b_{(l),i}} + w_u(\theta_i) \boldsymbol{\mu}_{b_{(u),i}}
\end{equation}
\begin{equation}
\label{eq:weightedvectors}
\boldsymbol{P}(\theta_i) = w_l(\theta_i) 
\begin{bmatrix}
\boldsymbol{p}_{b,1} &\hspace{-2mm}\boldsymbol{p}_{b,2} &\hspace{-2mm}\cdots &\hspace{-2mm}\boldsymbol{p}_{b,V}
\end{bmatrix} + w_u(\theta_i) \begin{bmatrix}
\boldsymbol{p}_{b+1,1} &\hspace{-2mm} \boldsymbol{p}_{b+1,2} &\hspace{-2mm} \cdots &\hspace{-2mm} \boldsymbol{p}_{b+1,V}
\end{bmatrix}
\end{equation}
The model produces a lower-dimensional representation of $\boldsymbol{x}_i$ as the coefficient vector $\boldsymbol{\beta}_i$.
This can be translated to a projection of $\boldsymbol{x}_i$ using $\boldsymbol{\mu}(\theta_i)+\boldsymbol{P}(\theta_i) \boldsymbol{\beta}_i$.}

\subsection{Energy Function}
\label{section:energy}
PPCA uses the minimization of an energy function $E(\cdot)$ to achieve a balance between having these projections fit the training examples well and reducing differences between adjacent bin endpoints' corresponding model components.
The energy $E(\cdot)$ contains three terms, a data fidelity term $E_\text{data}$ measuring the fitness to the input data, a smoothness term $E_\text{smo}$ that encourages smoothness for the means and principal vectors along the parameter, and an orthogonality term $E_\text{ortho.}$ that encourages the principal vectors at each bin endpoint to be orthogonal to each other.
\begin{equation}
\label{eq:energy}
E(\boldsymbol{\mu},\boldsymbol{p},\boldsymbol{\beta},\lambda_{m},\lambda_{v},\lambda_o) =
E_\text{data}(\boldsymbol{\mu},\boldsymbol{p},\boldsymbol{\beta}) +
E_\text{smo}(\boldsymbol{\mu},\boldsymbol{p},\lambda_{m},\lambda_{v}) +
E_\text{ortho.}(\boldsymbol{p},\lambda_o)
\end{equation}
\par{Equation \eqref{eq:energy} uses the vectors $\boldsymbol{\mu}$, $\boldsymbol{p}$, and
$\boldsymbol{\beta}$, which are stacked from vectors introduced earlier, as detailed in
Equation \eqref{eq:stackedvectors}.
The functions $\boldsymbol{\mu}(\theta_i)$ and $\boldsymbol{P}(\theta_i)$ can be derived from the vectors
$\boldsymbol{\mu}$ and $\boldsymbol{p}$, and the coefficient vectors $\boldsymbol{\beta}_i$ can be
extracted from $\boldsymbol{\beta}$.}
\begin{equation}
\label{eq:stackedvectors}
\boldsymbol{\mu} = \begin{bmatrix}
\boldsymbol{\mu}_1\\ \boldsymbol{\mu}_2\\ \vdots\\ \boldsymbol{\mu}_B
\end{bmatrix},
\boldsymbol{p} = \begin{bmatrix}
\boldsymbol{p}_{1,1}\\ \vdots\\ \boldsymbol{p}_{1,V}\\
\boldsymbol{p}_{2,1}\\ \vdots\\ \boldsymbol{p}_{B,V-1}\\ \boldsymbol{p}_{B,V}
\end{bmatrix},
\boldsymbol{\beta} = \begin{bmatrix}
\boldsymbol{\beta}_1\\ \boldsymbol{\beta}_2\\ \vdots\\ \boldsymbol{\beta}_n
\end{bmatrix}
\end{equation}
The first term from Equation \eqref{eq:energy} is the data term $E_\text{data}(\cdot)$,
which is the mean square approximation error over the training examples,

\begin{equation}
\label{eq:energydata}
E_\text{data}(\boldsymbol{\mu},\boldsymbol{p},\boldsymbol{\beta})=\frac{1}{n}
\displaystyle\sum_{i=1}^n{\|\boldsymbol{x}_i - \boldsymbol{\mu}(\theta_i) -
\boldsymbol{P}(\theta_i)\boldsymbol{\beta}_i\|^2}
\end{equation}
using the linear model $ \bmu(\theta_i) +\boldsymbol{P}(\theta_i)\bbeta_i$ based on paramater $\theta_i$ to approximate example $\bx_i$.

The second term, 
\begin{equation}
\label{eq:energysmoothness}
E_\text{smo}(\boldsymbol{\mu},\boldsymbol{p},\lambda_{m},\lambda_{v}) =
\frac{\lambda_{m}}{B-1}\sum_{b=1}^{B-1}{\|\boldsymbol{\mu}_b - \boldsymbol{\mu}_{b+1}\|^2} +
\frac{\lambda_{v}}{B-1}\sum_{b=1}^{B-1}{\sum_{v=1}^V{\|\boldsymbol{p}_{b,v}-
\boldsymbol{p}_{b+1,v}\|^2}}
\end{equation}
uses penalty coefficients $\lambda_{m}$ and $\lambda_{v}$ to ensure smooth functions for the
mean vectors and basis vectors, respectively.
Differences between corresponding elements for vectors relevant to two endpoints of the same bin are penalized.
Large values of $\lambda_{m}$ and $\lambda_{v}$ will enforce more smoothness in the representation,
at the expense of the projection error on the training set.

\par{PPCA's two smoothness penalty coefficients $\lambda_{m}$ and $\lambda_{v}$ force the model for an
observation to incorporate information from observations with similar observations:
those in its bin and those in the adjacent bin(s).
The amount of the information sharing depends on the differences in parameter values,
even for observations in the same bin.
The weighted pooling of information enforces a prior belief that observations with more similar values of a
parameter should be modeled in a more similar manner.
It enforces smoothness, but not monotonicity.
Ordinal trends can still be captured, but only locally.
This gives PPCA the ability to approximate more complicated smooth functions, though,
such as sinusoidal curves.
Like the prior beliefs in Bayesian models, PPCA's prior belief is more useful in the presence of limited training data, because the pooling of information can reduce overfitting.}
\par{The energy function also includes the orthogonality term $E_\text{ortho.}$, given in
Equation \eqref{eq:energyorthonormality}.
In Equation \eqref{eq:energyorthonormality}, the functions $\boldsymbol{1}_{(v=w)}$ are indicators for the condition $v=w$.
\begin{equation}
\label{eq:energyorthonormality}
E_\text{ortho.}(\boldsymbol{p},\lambda_o)=\lambda_o\sum_{b=1}^B{\sum_{v=1}^{V}{\sum_{w=v}^{V}
{\left(\langle\boldsymbol{p}_{b,v},\boldsymbol{p}_{b,w}\rangle - \boldsymbol{1}_{(v=w)}\right)^2}}}
\end{equation}
$E_\text{ortho.}(\cdot)$ encourages orthonormality in each bin endpoint's basis.
It penalizes differences from zero for dot products of pairs of vectors from the same bin, promoting orthogonality of each basis.
It also penalizes differences from one for the squared $\ell_2$ norm of each vector.}

\section{Learning a Parameterized Principal Component Analysis Model}
\label{section:ppcafitting}
\par{Because the energy function is composed of quadratic terms, we assume it to be locally convex.
We find a local minimum in the energy function using partial derivatives of the energy function with respect to
the stacked mean vector $\boldsymbol{\mu}$, the stacked basis vector $\boldsymbol{p}$,
and each observation's coefficient vector $\boldsymbol{\beta}_i$.
We either perform gradient descent or obtain the respective optimal component analytically by setting the derivatives to zero.}
\par{PPCA needs to choose optimal vectors $\boldsymbol{\mu}$, $\boldsymbol{p}$, and
$\boldsymbol{\beta}$, and a derivative-based method for one of the three requires knowing or estimating the
other two.
In PPCA, we optimize one at a time, holding the other two constant based on their most recent estimates.
After initialization, we run several cycles of optimizing the mean vectors, followed by the basis vectors,
and then the coefficient vectors.
We choose a pre-determined number of cycles $n_c$, and terminate the algorithm early if the algorithm is
deemed to have converged, based on the energy.
We store one previous iteration's estimates of the model components $\boldsymbol{\mu}$, $\boldsymbol{p}$, and $\boldsymbol{\beta}$, so these estimates can be treated as final if the energy increases.}

\subsection{Learning Mean Vectors}
\par{A closed-form solution for $\hat{\boldsymbol{\mu}}$, the PPCA estimate of $\boldsymbol{\mu}$, is displayed in Equation \eqref{eq:meanssolution}.
This uses the observation-specific matrix $\boldsymbol{W}_i$ from Equation \eqref{eq:weightmatrix}, which is made up of bin endpoint weights $w_{b,i}$.
The weight $w_{b,i}$ is equal to $w_l(\theta_i)$ if $b$ is the lower endpoint for observation $i$, $w_u(\theta_i)$ if $b$ is upper endpoint for observation $i$, and zero otherwise.
Equation \eqref{eq:meanssolution} also uses the weight-product matrix $\boldsymbol{C}_{(M),i}$ from Equation \eqref{eq:weightcrossproductmatrix} and the matrix $\boldsymbol{R}_{(M)}$ from Equation \eqref{eq:smoothnessmeansmatrixnomask}.
$\boldsymbol{R}_{(M)}$ has only three diagonals of non-zero elements, all of which are -1, 1, or 2.}
\begin{equation}
\label{eq:meanssolution}
\hat{\boldsymbol{\mu}}=\frac{1}{n}\left(\frac{1}{n}\displaystyle\sum_{i=1}^n
{\left[\boldsymbol{C}_{(M),i}\right]}+
\frac{\lambda_{m}}{B-1}\boldsymbol{R}_{(M)}\right)^{-1}\displaystyle\sum_{i=1}^n
{\left(\boldsymbol{W}_i^T\left[\boldsymbol{x}_i-\boldsymbol{P}(\theta_i)\boldsymbol{\beta}_i
\right]\right)}
\end{equation}
\begin{equation}
\label{eq:weightmatrix}
\boldsymbol{W}_i = \begin{bmatrix}
w_{1,i}I_K & w_{2,i}I_K & \cdots & w_{B,i}I_K
\end{bmatrix}
\end{equation}
\begin{equation}
\label{eq:weightcrossproductmatrix}
\boldsymbol{C}_{(M),i} = \begin{bmatrix}
w_{1,i}^2I_K & w_{1,i}w_{2,i}I_K & \cdots & w_{1,i}w_{B,i}I_K\\
w_{2,i}w_{1,i}I_K & w_{2,i}^2I_K & \cdots & w_{2,i}w_{B,i}I_K\\
\vdots & \vdots & \ddots & \vdots\\
w_{B,i}w_{1,i}I_K & w_{B,i}w_{2,i}I_K & \cdots & w_{B,i}^2I_K
\end{bmatrix}
\end{equation}

\begin{equation}
\label{eq:smoothnessmeansmatrixnomask}
\boldsymbol{R}_{(M)} = \begin{bmatrix}
I_K & -I_K & \boldsymbol{0}_{K \times K} & \cdots & \boldsymbol{0}_{K \times K} &
\boldsymbol{0}_{K \times K} & \boldsymbol{0}_{K \times K}\\
-I_K & 2I_K & -I_K & \cdots & \boldsymbol{0}_{K \times K} & \boldsymbol{0}_{K \times K} &
\boldsymbol{0}_{K \times K}\\
\boldsymbol{0}_{K \times K} & -I_K & 2I_K & \cdots & \boldsymbol{0}_{K \times K} &
\boldsymbol{0}_{K \times K} & \boldsymbol{0}_{K \times K}\\
\vdots & \vdots & \vdots & \ddots & \vdots & \vdots & \vdots\\
\boldsymbol{0}_{K \times K} & \boldsymbol{0}_{K \times K} & \boldsymbol{0}_{K \times K} & \cdots
& 2I_K & -I_K & \boldsymbol{0}_{K \times K}\\
\boldsymbol{0}_{K \times K} & \boldsymbol{0}_{K \times K} & \boldsymbol{0}_{K \times K} & \cdots
& -I_K & 2I_K & -I_K\\
\boldsymbol{0}_{K \times K} & \boldsymbol{0}_{K \times K} & \boldsymbol{0}_{K \times K} & \cdots
& \boldsymbol{0}_{K \times K} & -I_K & I_K
\end{bmatrix}
\end{equation}
\par{The use of a matrix inverse or linear system solution for $\hat{\boldsymbol{\mu}}$ is either impractically slow or inaccurate for high-dimensional data such as vectorized images.
For these data, we optimized the mean vectors using a gradient descent algorithm and the energy derivative from Equation \eqref{eq:meansderivative}.
\begin{equation}
\label{eq:meansderivative}
\frac{\partial E}{\partial\boldsymbol{\mu}} = -\frac{2}{n}\displaystyle\sum_{i=1}^n
{\left[\boldsymbol{C}_{(M),i}\left(\boldsymbol{y}_i-\boldsymbol{\mu}-
\boldsymbol{B}_i\boldsymbol{p}\right)\right]}+
\frac{2\lambda_{m}}{B-1}\boldsymbol{R}_{(M)}\boldsymbol{\mu}
\end{equation}
Equation \eqref{eq:meansderivative} uses the observation-specific coefficient matrices $\boldsymbol{B}_i$, which are defined using Equations \eqref{eq:coefficientmatrixrepeating} and \eqref{eq:coefficientmatrix}.
It also uses the stacked vectors $\boldsymbol{y}_i$, which stack $B$ identical copies of an observation $\boldsymbol{x}_i$.}
\begin{equation}
\label{eq:coefficientmatrixrepeating}
\boldsymbol{B}_i = \begin{bmatrix}
\boldsymbol{B}_{(B),i} & \boldsymbol{0}_{K \times KV} & \cdots & \boldsymbol{0}_{K \times KV}\\
\boldsymbol{0}_{K \times KV} & \boldsymbol{B}_{(B),i} & \cdots & \boldsymbol{0}_{K \times KV}\\
\vdots & \vdots & \ddots & \vdots\\
\boldsymbol{0}_{K \times KV} & \boldsymbol{0}_{K \times KV} & \cdots & \boldsymbol{B}_{(B),i}
\end{bmatrix}_{BK \times BKV}
\end{equation}
\begin{equation}
\label{eq:coefficientmatrix}
\boldsymbol{B}_{(B),i} = \begin{bmatrix}\beta_{i,1}I_K & \beta_{i,2}I_K & \cdots & \beta_{i,V}I_K\end{bmatrix}_{K \times KV}
\end{equation}

\subsection{Learning Basis Vectors}
\par{We only use gradient descent to optimize $\boldsymbol{p}$, because the presence of a dot product within a quadratic term creates a quartic term that prevents a closed-form solution.
The derivative is in Equation \eqref{eq:vectorsderivative}, and it relies on the $BKV$-length vectors $\boldsymbol{b}_i$, which stack products of the weights, coefficients, and residuals.}
\begin{multline}
\label{eq:vectorsderivative}
\frac{\partial E}{\partial\boldsymbol{p}} =
-\frac{2}{n}\displaystyle\sum_{i=1}^N{\boldsymbol{b}_i}
+\left(\frac{\lambda_{v}}{B-1}\boldsymbol{R}_{(V)}-4\lambda_{v}\right)\boldsymbol{p}+\\
2\lambda_o\sum_{b=1}^B{\sum_{v=1}^{V_b}{\sum_{w=v}^{V_b}{\left[\left(
\boldsymbol{T}_{b,v,w}+
\boldsymbol{T}_{b,w,v}\right)\boldsymbol{p}\boldsymbol{p}^T
\boldsymbol{T}_{b,w,v}\boldsymbol{p}\right]}}}
\end{multline}
\begin{equation}
\label{eq:weightcoeffientresidualvector}
\boldsymbol{b}_i = \begin{bmatrix}
w_{1,i}\beta_{i,1}\left[\boldsymbol{x}_i-\boldsymbol{\mu}(\theta_i)-
\boldsymbol{P}(\theta_i)\boldsymbol{\beta}_i\right] \\
w_{1,i}\beta_{i,2}\left[\boldsymbol{x}_i-\boldsymbol{\mu}(\theta_i)-
\boldsymbol{P}(\theta_i)\boldsymbol{\beta}_i\right] \\
\vdots \\
w_{B,i}\beta_{i,V-1}\left[\boldsymbol{x}_i-\boldsymbol{\mu}(\theta_i)-
\boldsymbol{P}(\theta_i)\boldsymbol{\beta}_i\right] \\
w_{B,i}\beta_{i,V}\left[\boldsymbol{x}_i-\boldsymbol{\mu}(\theta_i)-
\boldsymbol{P}(\theta_i)\boldsymbol{\beta}_i\right]
\end{bmatrix}
\end{equation}
\par{Equation \eqref{eq:vectorsderivative} also uses the transition-like matrix $\boldsymbol{T}_{b,v,w}$ from Equation \eqref{eq:transitionmatrix} and the bin-comparison matrix $\boldsymbol{R}_{(V)}$ from Equation \eqref{eq:smoothnessvectorsmatrixnomask}.
$\boldsymbol{T}_{b,v,w}$, if multiplied by $\boldsymbol{p}$, will zero out all $\boldsymbol{p}_{b,w^*}$ except for $\boldsymbol{p}_{b,w}$, which gets moved to the appropriate spot for $\boldsymbol{p}_{b,v}$.
In its definition, the functions $\boldsymbol{1}_{(\cdot)}$ are indicator functions for the events within the parentheses.
$\boldsymbol{R}_{(V)}$ is a larger version of the matrix $\boldsymbol{R}_{(M)}$ used for the means.}
\begin{equation}
\label{eq:transitionmatrix}
\boldsymbol{T}_{b,v,w}\hspace{-1mm} = \hspace{-1.5mm}\begin{bmatrix}
\boldsymbol{1}_{(b=1 \cap v=1 \cap w=1)}I_K &\hspace{-2mm}\boldsymbol{1}_{(b=1 \cap v=1 \cap w=2)}I_K\hspace{-2mm} &
\cdots & \boldsymbol{0}_{K \times K}\\
\boldsymbol{1}_{(b=1 \cap v=2 \cap w=1)}I_K &\hspace{-2mm}\boldsymbol{1}_{(b=1 \cap v=2 \cap w=2)}I_K\hspace{-2mm} &
\cdots & \boldsymbol{0}_{K \times K}\\
\vdots & \vdots & \ddots & \vdots\\
\boldsymbol{0}_{K \times K} & \boldsymbol{0}_{K \times K} & \cdots &
\hspace{-2mm}\boldsymbol{1}_{(b=B \cap v=V \cap w=V)}I_K
\end{bmatrix}
\end{equation}
\begin{equation}
\label{eq:smoothnessvectorsmatrixnomask}
\boldsymbol{R}_{(V)} = \begin{bmatrix}
I_{KV} & -I_{KV} & \cdots & \boldsymbol{0}_{KV \times KV} & \boldsymbol{0}_{KV \times KV}\\
-I_{KV} & 2I_{KV} & \cdots & \boldsymbol{0}_{KV \times KV} & \boldsymbol{0}_{KV \times KV}\\
\boldsymbol{0}_{KV \times KV} & -I_{KV} & \cdots & \boldsymbol{0}_{KV \times KV} &
\boldsymbol{0}_{KV \times KV}\\
\vdots & \vdots & \ddots & \vdots & \vdots\\
\boldsymbol{0}_{KV \times KV} & \boldsymbol{0}_{KV \times KV} & \cdots & -I_{KV} &
\boldsymbol{0}_{KV \times KV}\\
\boldsymbol{0}_{KV \times KV} & \boldsymbol{0}_{KV \times KV} & \cdots & 2I_{KV} & -I_{KV}\\
\boldsymbol{0}_{KV \times KV} & \boldsymbol{0}_{KV \times KV} & \cdots & -I_{KV} & I_{KV}
\end{bmatrix}
\end{equation}
\par{The gradient descent algorithm has a soft constraint for orthonormal bases, but we implement a hard constraint for normality as well.
After the gradient descent algorithm for $\boldsymbol{p}$ completes, we rescale each basis vector $\boldsymbol{p}_{b,v}$ to have a unit norm. We cannot similarly force orthogonality without undoing the gradient descent algorithm's attempts to enforce smoothness.}

\subsection{Learning Coefficient Vectors}
\label{section:coefficients}
\par{If one differentiates the energy function with respect to a single observation's coefficient vector $\boldsymbol{\beta}_i$ and sets this derivative equal to the zero vector, one can obtain the estimate $\hat{\boldsymbol{\beta}}_i$ below for a coefficient vector $\boldsymbol{\beta}_i$.
\begin{equation}
\label{eq:coefficientssolution}
\hat{\boldsymbol{\beta}}_i = \left[ \boldsymbol{P}(\theta_i) \right]^{-1} \left[ \boldsymbol{x}_i - \boldsymbol{\mu}(\theta_i) \right]
\end{equation}
This inverse is applied to a much smaller matrix than that inverted to find $\hat{\boldsymbol{\mu}}$, so we use a linear system solution to obtain $\hat{\boldsymbol{\beta}}_i$, even with high-dimensional data.}

\subsection{Initialization}
\label{section:initialization}
\par{PPCA finds an appropriate local minimum within the energy function, so an appropriate initialization is
important for finding a local minimum that can perform similarly to the global minimum.
We initialize PPCA using a procedure similar to IPCA, which runs PCA on groups made by binning the parameter $\theta$.
We calculate initial mean vectors $\hat{\boldsymbol{\mu}}_{(0),b}$ using Equation \eqref{eq:initialmeans},
which is like a weighted version of the mean calculation from IPCA.}

\begin{figure}
\centering
\includegraphics[scale=0.6]{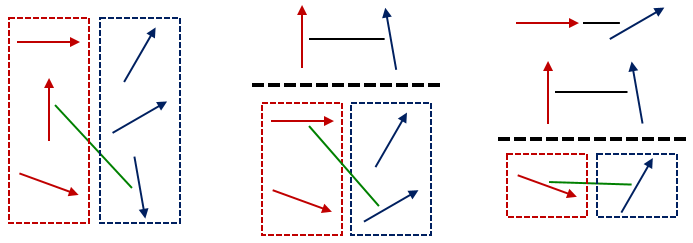}
\vskip -4mm
\caption{Example reordering and sign change of initial basis vectors}
\label{fig:vectorreordering}
\vspace{-3mm}
\end{figure}

\begin{equation}
\label{eq:initialmeans}
\hat{\boldsymbol{\mu}}_{(0),b}=\frac{\sum_{i=1}^n{w_{b,i}\boldsymbol{x}_i}}{\sum_{i=1}^n{w_{b,i}}}
\end{equation}
\par{To find the initial basis vectors, we choose overlapping subsets of the observations $\boldsymbol{x}_i$
and assign one subset to each bin endpoint $b$.
The included observations are all with weight values $w_{b,i}$ above a given threshold such as 0.001.
We run PCA on each of these subsets, except that we use the means $\hat{\boldsymbol{\mu}}_{(0),b}$
instead of recalculating the means based on the subsets of $\boldsymbol{x}_i$.
We then reorder these PCA basis vectors to promote smoothness, using a greedy algorithm.
One can start with the first bin endpoint's basis as the first reference basis, and reorder the bases from the second until the last bin endpoint.
Alternatively, one can make the last bin endpoint's basis the first reference basis, and reorder the bases from the second-to-last until the first bin endpoint.}
\par{For each pair of bin endpoints, one first calculates the absolute values of the dot products between each pair of basis vectors using one from each endpoint.
The two vectors with the highest absolute value of the dot product are paired, and the sign is inverted for the vector from the basis to reorder if the dot product is negative.
This procedure continues, each time only using vectors that are not in any pairs, until all vectors in the reference basis have been paired.
If any vectors remain in the basis to reorder, they are assigned to any unused locations.
The basis just reordered then becomes the reference basis, the next basis in the order is assigned to be
reordered, and the procedure continues until all bases except the original reference have been reordered.
The coefficients can then be initialized from the initial mean and basis vectors using Equation \eqref{eq:coefficientssolution}.}

\subsection{Putting it all together}
\label{section:algorithm}
The complete PPCA training algorithm is summarized in Algorithm \ref{alg:ppca}.
\begin{algorithm}[htb]
\caption{PPCA training algorithm}\label{alg:ppca}
\begin{algorithmic}[1]
%\For {$i=1$ to $n$}
	\State {set $w_l(\theta_i)$ and $w_u(\theta_i), i=\overline{1,n}$ using Equation \eqref{eq:weights}}
%\EndFor
\For {$b=1$ to $B$}
	\State {initialize $\hat{\boldsymbol{\mu}_b}$ using Equation \eqref{eq:initialmeans}}
	\State {initialize vectors $\hat{\boldsymbol{p}}_{b,v}$ using PCA on examples with $w_{b,i} > \epsilon$}
	\State {rearrange vectors $\hat{\boldsymbol{p}}_{b,v}$ for same $b$ and switch signs if necessary}
\EndFor
%\For {$i=1$ to $n$}
	\State {initialize $\hat{\boldsymbol{\beta}}_i,  i=\overline{1,n}$ using Equation \eqref{eq:coefficientssolution}}
%\EndFor
\State {find $E_0$ using Equation \eqref{eq:energy}}
\For {$c=1$ to $n_c$}
	\State {update $\hat{\boldsymbol{\mu}}$ using Equation \eqref{eq:meanssolution} or gradient descent
	with Equation \eqref{eq:meansderivative}}
	\State {update $\hat{\boldsymbol{p}}$ using gradient descent with Equation \eqref{eq:vectorsderivative}}
	%\For {$i=1$ to $n$}
		\State {update $\hat{\boldsymbol{\beta}}_i,  i=\overline{1,n}$ using Equation \eqref{eq:coefficientssolution}}
	%\EndFor
	\State {find $E_c$ using Equation \eqref{eq:energy}}
	\If {$E_c > E_{c-1}$}
		\State {\textbf{break}}
	\EndIf
\EndFor
\end{algorithmic}
\end{algorithm}

\subsection{Tuning of Parameters}
\label{section:parametertuning}
\par{The energy must be tracked, so one can use its path to choose the number of overall cycles $n_c$, the learning rates ($\alpha_m$ for means, $\alpha_v$ for bases), the number of iterations with those learning rates ($n_m$ for means, $n_v$ for bases), and the non-orthonormality penalty coefficient $\lambda_o$.
If one wants to choose appropriate smoothness penalty coefficients ($\lambda_{m}$ for means, $\lambda_{v}$ for bases), then one should tune them using a validation set selected randomly from the training examples.
Typically, $\lambda_o$ should be much larger than $\lambda_{v}$.
However, $\alpha_v$ must decrease as $\lambda_o$ increases, so an excessively large $\lambda_o$ leads to unnecessary increases in run-time.}

\section{Modifications and Generalizations for Real Applications}
\label{section:ppcageneralizations}
This section details two modifications for generalizations that allow dimensions to vary with the PPCA parameter.
The first is for the dimension of the manifold, and the second is for the observations.

\subsection{Generalization to Varied Manifold Dimension}
\label{section:variednumbervectors}
\par{For some applications, the manifold dimension can vary with the parameter $\theta$.
%In Section \ref{section:lymphnodes}, we present such a case, in which higher values of $\theta$ are thought to
%require a more complex representation (with a larger coefficient vector $\boldsymbol{\beta}_i$).
In this case, each bin endpoint $b$ would have $V_b$ basis vectors, and $V$ would be set to the largest $V_b$.
One would still allocate $V$ basis vectors in $\boldsymbol{p}$ for each bin endpoint, but one would set $\boldsymbol{p}_{b,v}$ to be a zero vector if $v>V_b$.}

\begin{multline}
\label{eq:energysmoothnessvariednumbervectors}
E_\text{smo}(\boldsymbol{\mu},\boldsymbol{p},\lambda_{m},\lambda_{v}) =
\frac{\lambda_{m}}{B-1}\sum_{b=1}^{B-1}{\|\boldsymbol{\mu}_b - \boldsymbol{\mu}_{b+1}\|^2} +\\
\frac{\lambda_{v}}{B-1}\sum_{b=1}^{B-1}{\sum_{v=1}^{\min\left(V_b,V_{b+1}\right)}
{\|\boldsymbol{p}_{b,v}-\boldsymbol{p}_{b+1,v}\|^2}}
\end{multline}
\par{If one has differently-sized bases, the energy component $E_\text{smo}$ needs to follow Equation
\eqref{eq:energysmoothnessvariednumbervectors} instead of Equation \eqref{eq:energysmoothness}.
Also, the energy component $E_\text{ortho.}$ needs to follow Equation
\eqref{eq:energyorthonormalityvariednumbervectors} instead of Equation \eqref{eq:energyorthonormality}.
\begin{equation}
\label{eq:energyorthonormalityvariednumbervectors}
E_\text{ortho.}(\boldsymbol{p},\lambda_o)=\lambda_o\sum_{b=1}^B{\sum_{v=1}^{V_b}
{\sum_{w=v}^{V_b}{\left(\langle\boldsymbol{p}_{b,v},\boldsymbol{p}_{b,w}\rangle -
\boldsymbol{1}_{(v=w)}\right)^2}}}
\end{equation}
The only three changes to these two equations are to the upper boundaries of summations.
In Equation \eqref{eq:energysmoothnessvariednumbervectors}, the third summation ends at
$\min\left(V_b,V_{b+1}\right)$ rather than at $V$.
This is intended so PPCA only enforces similarity between the corresponding basis vectors for adjacent bin
endpoints if the basis vectors exist for both.
In Equation \eqref{eq:energyorthonormalityvariednumbervectors}, the second and third summations end at
$V_b$ instead of at $V$.
This is because there are no vectors beyond vector $V_b$ upon which to enforce orthonormality.}
\begin{figure}[htb]
\centering
\includegraphics[scale=0.55]{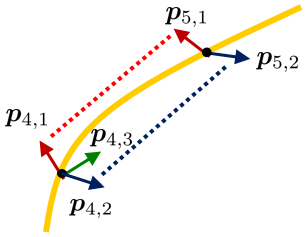}
\vskip -4mm
\caption{Example of a varying number of basis vectors, where vector $\bp_{4,3}$ has no correspondent in bin $5$.}
\label{fig:vectorcorrespondence}
\vspace{-3mm}
\end{figure}

\par{In Section \ref{section:initialization}, we detailed a procedure of rearranging initial basis vectors produced by PCA.
If the number of basis vectors is either non-decreasing or non-increasing with respect to the bin endpoint number, then this procedure still works, with one modification.
If all bin endpoints use $V$ basis vectors, the user has the choice of reordering from the second until the last bin endpoint, or from the second-to-last until the first bin endpoint.
However, if the first bin endpoint's basis is smaller than the last bin endpoint's basis, the reordering procedure must go from the second basis to the last.
If the last bin endpoint's basis is smaller than the first bin endpoint's basis, the reordering procedure must go from the second-to-last basis to the first.
If the number of basis vectors both increases and decreases when going from the first to the last bin endpoint, then the reordering must be done more manually.}

\subsection{Generalization to Varying Manifold Ambient Space}
\label{section:mask}
\par{Applications also exist in which the manifold ambient space varies with the parameter $\theta$.
Section \ref{section:yaw} demonstrates an example of this sort, using face images.
In these data, certain pixels may be considered outside the face shape for a given face.
Like the observations, the mean and basis vectors for bin endpoint $b$ may not use all $K$ elements.
As shown in Figure \ref{fig:maskcorrespondence}, we want the mean vectors
$\boldsymbol{\mu}_b=(\mu_{b,1},\ldots,\mu_{b,K})^T$ to have similarity enforced between elements
$\mu_{b,k}$ and $\mu_{b+1,k}$ only if element $k$ is relevant for both bin endpoint $b$ and bin endpoint $b+1$.
So, we create the indicator variables $m_{b,k}$ which equal one if element $k$ is included for bin endpoint $b$,
and zero otherwise.
From these, we can construct matrices $\boldsymbol{M}_{(1),b}$ that can adjust mean vectors
$\boldsymbol{\mu}_b$ or basis vectors $\boldsymbol{p}_{b,v}$, setting unused elements to zero.
We also construct matrices $\boldsymbol{M}_{(R1),b}$ as shown in Equation
\eqref{eq:maskvectorsmatrix1endpoint}, which can similarly adjust larger vectors.
\begin{equation}
\label{eq:maskmeansmatrix1endpoint}
\boldsymbol{M}_{(1),b} = \begin{bmatrix}
m_{b,1} & 0 & \cdots & 0\\
0 & m_{b,2} & \cdots & 0\\
\vdots & \vdots & \ddots & \vdots\\
0 & 0 & \cdots & m_{b,K}\\
\end{bmatrix}_{K \times K}
\end{equation}
\begin{equation}
\label{eq:maskvectorsmatrix1endpoint}
\boldsymbol{M}_{(R1),b} = \begin{bmatrix}
\boldsymbol{M}_{(1),b} & \boldsymbol{0}_{K \times K} & \cdots & \boldsymbol{0}_{K \times K}\\
\boldsymbol{0}_{K \times K} & \boldsymbol{M}_{(1),b} & \cdots & \boldsymbol{0}_{K \times K}\\
\vdots & \vdots & \ddots & \vdots\\
\boldsymbol{0}_{K \times K} & \boldsymbol{0}_{K \times K} & \cdots & \boldsymbol{M}_{(1),b}\\
\end{bmatrix}_{KV \times KV}
\end{equation}

\begin{figure}[htb]
\centering
\includegraphics[scale=0.45]{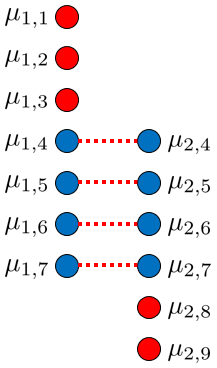}
\vskip -3mm
\caption{Example of mean vectors from two endpoints of same bin, in the scenario of a varying ambient space of the manifold}
\label{fig:maskcorrespondence}
\vspace{-3mm}
\end{figure}

For each bin endpoint $b$, one would then calculate
$\boldsymbol{M}_{(2),b}=\boldsymbol{M}_{(1),b}\boldsymbol{M}_{(1),b+1}$ and
$\boldsymbol{M}_{(R2),b}=\boldsymbol{M}_{(R1),b}\boldsymbol{M}_{(R1),b+1}$.
$\boldsymbol{M}_{(2),b}$ can then be used to adjust the energy component $E_\text{smo}$ as
shown in Equation \eqref{eq:energysmoothnesswithmask}.
The only adjustments made relative to Equation \eqref{eq:energysmoothness} are two additions of $\boldsymbol{M}_{(2),b}$.}
\begin{multline}
\label{eq:energysmoothnesswithmask}
E_\text{smo}(\boldsymbol{\mu},\boldsymbol{p},\lambda_{m},\lambda_{v}) =
\frac{\lambda_{m}}{B-1}\sum_{b=1}^{B-1}{\|\boldsymbol{M}_{(2),b}\left(\boldsymbol{\mu}_b -
\boldsymbol{\mu}_{b+1}\right)\|^2} +\\
\frac{\lambda_{v}}{B-1}\sum_{b=1}^{B-1}{\sum_{v=1}^V{\|\boldsymbol{M}_{(2),b}
\left(\boldsymbol{p}_{b,v}-\boldsymbol{p}_{b+1,v}\right)\|^2}}
\end{multline}
\par{The matrices $\boldsymbol{R}_{(M)}$ and $\boldsymbol{R}_{(V)}$ from Equations \eqref{eq:meanssolution}, \eqref{eq:meansderivative}, and \eqref{eq:vectorsderivative} must be modified as well.
These each still have three diagonals that can have non-zero elements, but these diagonals incorporate the
indicator variables $m_{b,k}$ and thus can have zeros.
The modified versions, shown in Equations \eqref{eq:smoothnessmeansmatrixwithmask} and
\eqref{eq:smoothnessvectorsmatrixwithmask}, only differ from Equations
\eqref{eq:smoothnessmeansmatrixnomask} and \eqref{eq:smoothnessvectorsmatrixnomask}
by including $\boldsymbol{M}_{(2),b}$ and $\boldsymbol{M}_{(R2),b}$, respectively, instead
of identity matrices of the same size.}
\begin{equation}
\label{eq:smoothnessmeansmatrixwithmask}
\boldsymbol{R}_{(M)} \hspace{-1mm}= \hspace{-1mm}\begin{bmatrix}
\boldsymbol{M}_{(2),1} & -\boldsymbol{M}_{(2),1} & \cdots & \boldsymbol{0}_{K \times K} & \boldsymbol{0}_{K \times K}\\
\hspace{-0.5mm}-\boldsymbol{M}_{(2),1} & \hspace{-1mm}\boldsymbol{M}_{(2),1} \hspace{-1mm}+ \hspace{-1mm}\boldsymbol{M}_{(2),2}\hspace{-2mm} & \cdots & \boldsymbol{0}_{K \times K} &
\boldsymbol{0}_{K \times K}\\
\boldsymbol{0}_{K \times K} & -\boldsymbol{M}_{(3),2} & \cdots & \boldsymbol{0}_{K \times K} &
\boldsymbol{0}_{K \times K}\\
\vdots & \vdots & \ddots & \vdots & \vdots\\
\boldsymbol{0}_{K \times K} & \boldsymbol{0}_{K \times K} & \cdots & -\boldsymbol{M}_{(2),B-2} &
\boldsymbol{0}_{K \times K}\\
\boldsymbol{0}_{K \times K} & \boldsymbol{0}_{K \times K} & \cdots & \hspace{-2mm}\boldsymbol{M}_{(2),B-2} \hspace{-1mm}+\hspace{-1mm} \boldsymbol{M}_{(2),B-1} \hspace{-1mm}& -\boldsymbol{M}_{(2),B-1}\hspace{-0.5mm}\\
\boldsymbol{0}_{K \times K} & \boldsymbol{0}_{K \times K} & \cdots & -\boldsymbol{M}_{(2),B-1} &
\boldsymbol{M}_{(2),B-1}
\end{bmatrix}
\end{equation}
\begin{equation}
\label{eq:smoothnessvectorsmatrixwithmask}
\boldsymbol{R}_{(V)} = \begin{bmatrix}
\boldsymbol{M}_{(R2),1} & -\boldsymbol{M}_{(R2),1} & \cdots & \boldsymbol{0}_{KV \times KV}\\
-\boldsymbol{M}_{(R2),1} & \boldsymbol{M}_{(R2),1} + \boldsymbol{M}_{(R2),2} & \cdots &
\boldsymbol{0}_{KV \times KV}\\
\boldsymbol{0}_{KV \times KV} & -\boldsymbol{M}_{(R2),2} & \cdots & \boldsymbol{0}_{KV \times KV}\\
\vdots & \vdots & \ddots & \vdots\\
\boldsymbol{0}_{KV \times KV} & \boldsymbol{0}_{KV \times KV} & \cdots &
\boldsymbol{0}_{KV \times KV}\\
\boldsymbol{0}_{KV \times KV} & \boldsymbol{0}_{KV \times KV} & \cdots & -\boldsymbol{M}_{(R2),B-1}\\
\boldsymbol{0}_{KV \times KV} & \boldsymbol{0}_{KV \times KV} & \cdots & \boldsymbol{M}_{(R2),B-1}
\end{bmatrix}
\end{equation}

\section{Experiments}
\label{section:experiments}
\par{We evaluated PPCA on four datasets. One had data created from known parameters, so that we could evaluate how well can PPCA recover these parameters. The other three were for applications of PPCA to real data: shapes for lymph nodes of varied sizes, appearances for faces of varied blurriness, and appearances for faces of varied yaw rotation.}

\noindent {\bf Parameter tuning.} In these experiments the learning rates $\alpha_m,\alpha_v$ were chosen as  the combination $(\alpha_m,\alpha_v) \in \{10^{-2},...,10^{-6}\}^2$ that obtained the smallest value of the energy function \eqref{eq:energy}.

\subsection{Simulation Experiments}
\label{section:artificialdata}
\par{First, we tested PPCA's ability to recover a true model, using three-dimensional data created from known mean and basis vectors.
These were based on smooth functions of a known parameter $\theta$, defined on the range from 0 to 360.
We used 45 observations with $\theta=4,12,20,\ldots,356$.
The data were based on two basis vectors and on coefficients drawn independently from a $U(-1,1)$ distribution.
We also added random noise to each element, using a $U(-1.5,1.5)$ distribution.
The formulas for the true mean vectors $\boldsymbol{\mu}(\theta)$ and true basis vectors $\boldsymbol{p}_1(\theta)$ and $\boldsymbol{p}_2(\theta)$ were as follows.}
\begin{equation}
\label{eq:artificialmeans}
\boldsymbol{\mu}(\theta)=\left \{\sin{\left(\frac{7\pi\theta}{720}\right)}, -\frac{91\theta}{1800}+8,
\sin{\left(\frac{7\pi\theta}{576}+0.6\right)}\right \}^T
\end{equation}
\begin{equation}
\label{eq:artificialvector1}
\boldsymbol{p}_1(\theta)=\left \{ \sin{\left(\frac{7\pi\theta}{1080}+0.4\right)},
\tan{\left(\frac{7\pi\theta}{4860}-0.8\right)}, \frac{49\theta}{1800}-1.1 \right \}^T
\end{equation}
\begin{equation}
\label{eq:artificialvector2}
\boldsymbol{p}_2(\theta)=\left \{ \cos{\left(\frac{7\pi\theta}{972}\right)},
\cos{\left(\frac{7\pi\theta}{576}-0.4\right)}, \frac{7\theta}{600}+1.4 \right \}^T
\end{equation}

\begin{figure}[htb]
\centering
\includegraphics[scale=0.4]{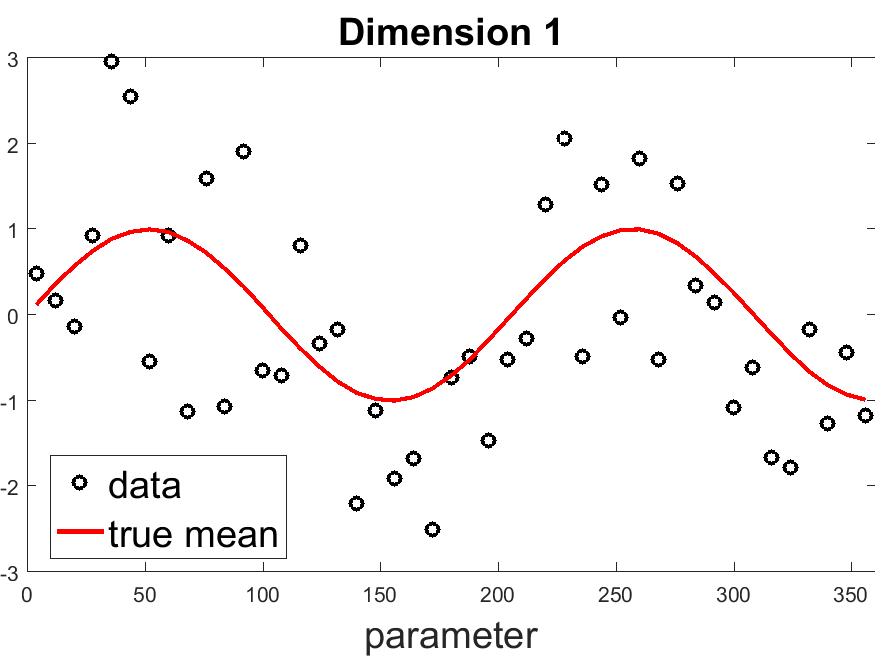}
\includegraphics[scale=0.4]{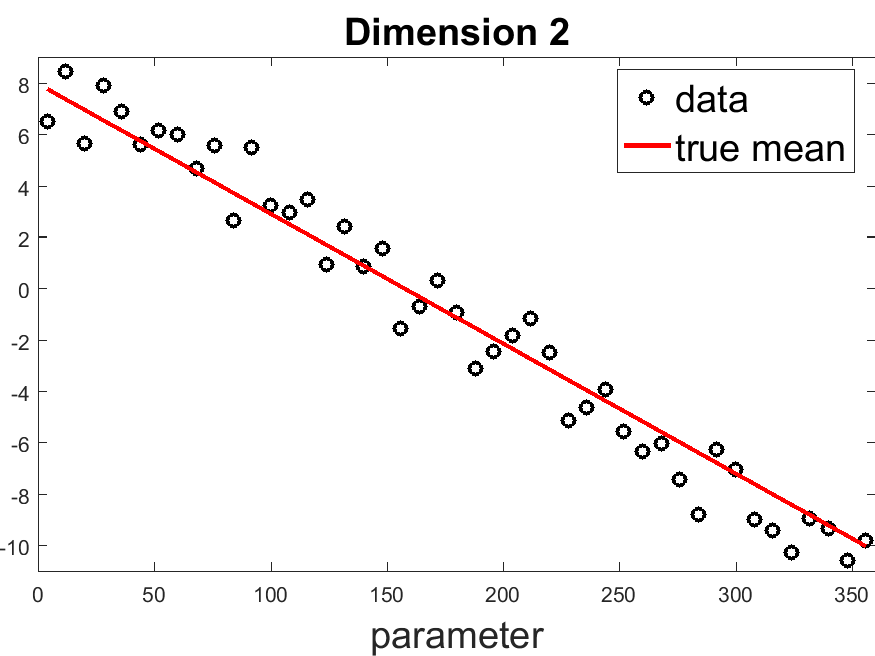}
\includegraphics[scale=0.4]{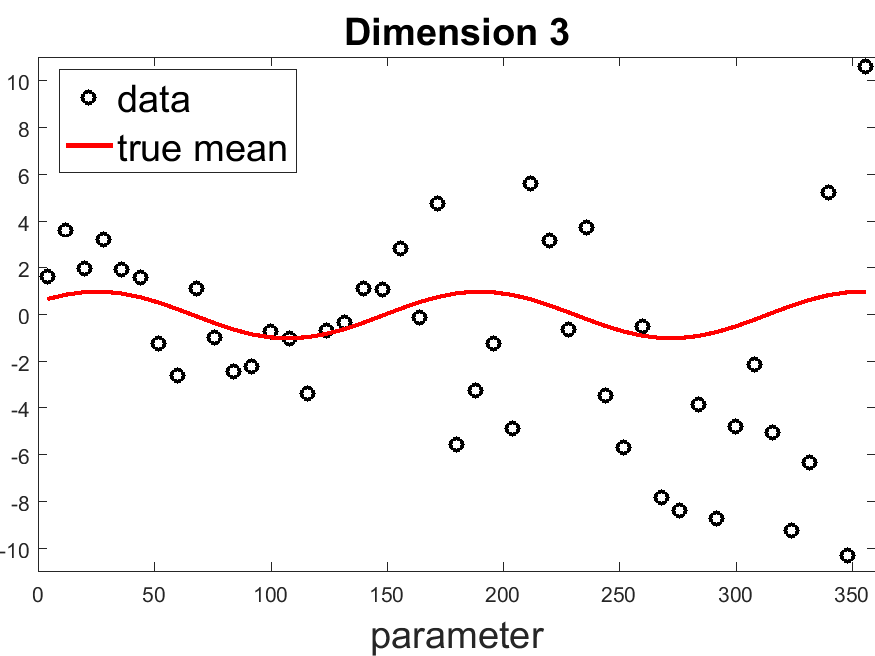}
\vskip -6mm
\caption{Artificial data with its three dimensions and the true mean function that was used to generate it.}
\label{fig:artificialdata}
\vspace{-2mm}
\end{figure}

\par{We divided the acceptable parameter range into 14 equally-sized bins.
Because the data and model were small (three dimensions and two basis vectors), we could use the analytical solution to calculate the mean vectors and only needed gradient descent for the bases.
We tested various smoothness penalty coefficients, but always used $\lambda_o=20$, $n_c=1000$, and $n_v=500$.}

\begin{figure}[tb]
\centering
\includegraphics[scale=0.4]{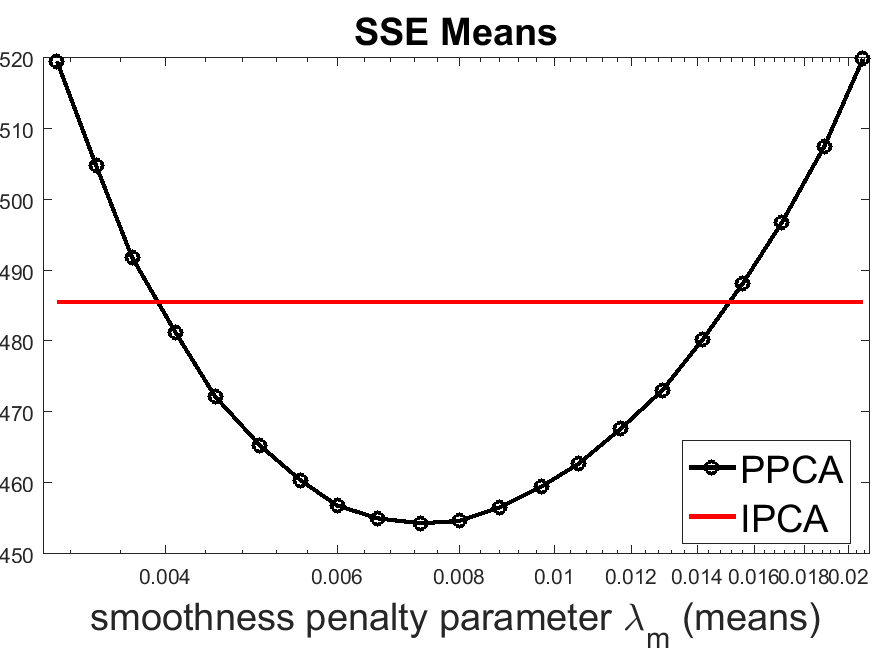}
\includegraphics[scale=0.4]{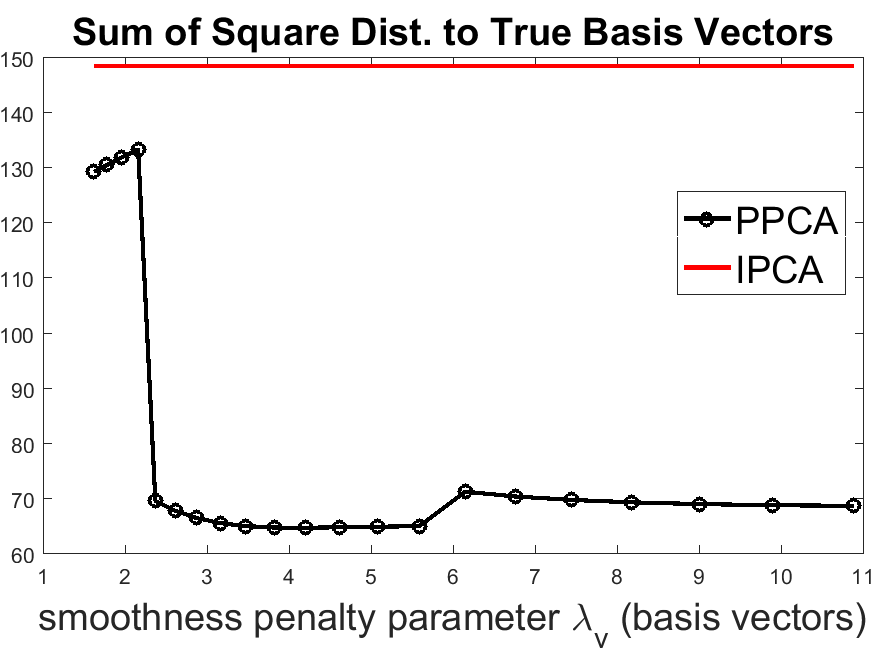}
\vskip -3mm
\caption{Parameter robustness experiment on the artificial data. Left: dependence of the estimated means on the smoothness penalty parameter $\lambda_{m}$, computed as the SSE to the true means. Right: dependence of the estimated vectors on the smoothness penalty parameter $\lambda_{v}$, computed as the sum of squared distances to the true basis vectors.}
\label{fig:errorartificialdata}
\vspace{-3mm}
\end{figure}

Figure \ref{fig:errorartificialdata} shows the sum of squared $\ell_2$ norms for the error in PPCA's and IPCA's estimates of the mean vector, compared to the true mean vectors $\boldsymbol{\mu}(\theta)$.
This uses various $\lambda_{m}$ but fixes $\lambda_{v}=4.2$.
For the bases, we could not use a simple $\ell_2$ error because a proper recovery could have the same linear subspace, but different vectors and coefficients.
We instead measured the $\ell_2$ norms of the normal vectors from the planes created by the recovered basis vectors to each of the true basis vectors.
Figure \ref{fig:errorartificialdata} shows the sums (across the observations and two true vectors) of these squared $\ell_2$ distances.
This uses various $\lambda_{v}$ but fixes $\lambda_{m}=0.008$.

\subsection{Lymph Node Segmentation}
\label{section:lymphnodes}
\par{Lymph nodes are organs that are part of the circulatory system and the immune system, which are important for the diagnosis and treatment of cancer and other medical conditions.
For cancer patients, one may want a segmentation for targeting radiation or for volume estimates.
Lymph node sizes generally increase with the onset of cancer, and decrease as treatment succeeds, so volume estimates are used to assess the efficacy of treatment.
Generally, radiologists use 3D computed tomography (CT) to assess lymph nodes, and lymph nodes tend to have spherical, elliptical, and bean-like shapes.
However, their shape can become more complicated as their size increases.
Barbu et al. (2012) demonstrated a model for representing lymph nodes by the lengths of radii, which extend in 162 pre-determined directions from the lymph node's center \cite{barbu2012automatic}.
We reduce this 162-dimensional representation of a lymph node's shape even further, using PPCA and IPCA with 6 dimensions.}
\par{We had a dataset available, in which an experienced radiologist had manually segmented 592 lymph nodes
from various patients treated at the National Institutes of Health Clinical Center.
We used only the 397 lymph nodes for which the 162-dimensional model was most appropriate.
We eliminated 182 lymph nodes which were part of conglomerates of lymph nodes,
and 16 for which the radial model's S\o rensen-Dice coefficient was less than 0.8.
We then randomly selected 79 lymph nodes to be in the test set,
and assigned the remaining 318 lymph nodes to the training set.}
\par{The parameter of interest for this application was the lymph node's diameter, because lymph nodes' sizes
are related to the types of shapes they can take.
We used an estimated diameter based on the 162 modeled radii.
We divided the lymph nodes into bins of 6-12, 12-18, 18-24, and 24-43 millimeters.
The dataset had 55, 156, 77, and 30 training examples per bin and 16, 39, 17, and 7 test examples per bin,
in increasing order of bin.
We used 6 basis vectors, for the IPCA bins and for the PPCA bin endpoints.
%These were chosen to allow more shape complexity for the larger lymph nodes,
%and to ensure that PPCA and IPCA could be compared fairly, with the same average dimensions used per bin.
We also evaluated PCA with 6 principal vectors and Sparse PCA \cite{zou2006sparse}\footnote{Using the implementation from \url{http://www2.imm.dtu.dk/projects/spasm/}} (SPCA) with 30 principal vectors with 50 nonzero entries in each vector.
}
\par{We evaluated all methods after fitting the models using different numbers of training examples per bin,
from 2 to 30. 
These smaller training sets were chosen randomly from the full training set. 
For PCA and SPCA we trained on the same examples as the other methods, but without using the bin information.
All smaller training sets were subsets of the larger training sets,
to ensure a more valid comparison of the effect of the training set size.
Each time, we calculated the root mean squared error (RMSE) of the approximation of each lymph node's
162-dimensional vector of radii, and then found the mean RMSE across the lymph nodes of the training or test set.
For PPCA, we used gradient descent algorithms for both the mean and basis vectors.
We used $\lambda_{m}=0.007$, $\lambda_{v}$ = 30,000, $\lambda_o=10^7$, $n_c=200$, $n_m=100$,
$n_v=100$.}
\begin{figure}
\centering
\includegraphics[scale=0.4]{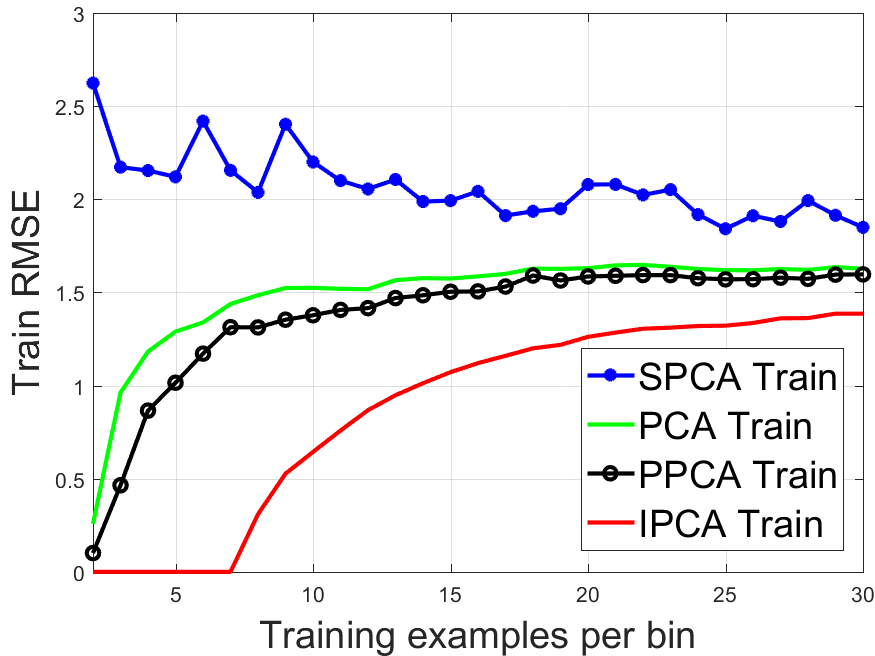}
\includegraphics[scale=0.4]{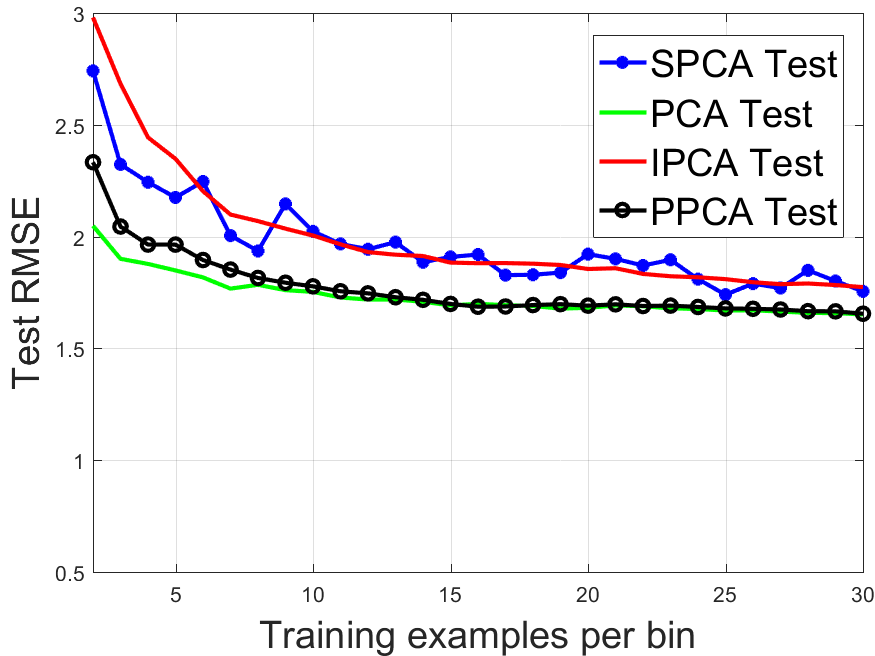}
\vskip -4mm
\caption{Mean RMSE for projection of radial representation of lymph nodes, evaluated on training sets (left) and test sets (right) using varied numbers of training examples.}
\label{fig:rmselymphnodes}
\vspace{-3mm}
\end{figure}
\par{Figure \ref{fig:rmselymphnodes} shows that IPCA overfits the data compared to PPCA, particularly for smaller training sets.
For all tested sizes, PPCA had noticeably higher error than IPCA when projecting the training set, but noticeably lower error than IPCA did when projecting the test set. The SPCA test RMSEs vary quite a lot around the RMSE of the IPCA. Observe that SPCA actually underfits the data since the training RMSEs are high and comparable to the test RMSEs.
PCA does a very good job at approximating this data, which means that probably the manifold is close to linear in this case. Some key results are also summarized in Table \ref{tab:rmseln}.
}
\begin{table}[htb]
\vspace{-3mm}
\small
\begin{center}
\caption{Summary of RMSE results for the lymph node data.}\label{tab:rmseln}
%\vskip -1mm
\begin{tabular}{|c|cccc|cccc|}
\hline
Training examples    &\multicolumn{4}{c}{Train RMSE} &\multicolumn{4}{|c|}{Test RMSE} \\
per bin  &PCA  &IPCA &PPCA &SPCA &PCA &IPCA &PPCA &SPCA\\
\hline
2   &0.258 &0.000	&0.101 &2.624 &2.048 &2.980	&2.333 &2.741\\
\hline
10 &1.525 &0.644	&1.378 &2.200 &1.753 &2.005	&1.777 &2.023\\
\hline
20 &1.630 &1.262	&1.586 &2.079 &1.681 &1.855	&1.691 &1.921\\
\hline
30 &1.627 &1.386	&1.597 &1.850 &1.652 &1.774	&1.655 &1.754\\
\hline
\end{tabular}
\end{center}
\vspace{-6mm}
\end{table}

\subsection{Facial Images with Blur}
\label{section:blur}
\par{Modeling images of human faces in photos or video frames is useful for creating novel images that satisfy the constraints of realistic faces, such as for animation.
It can also modify a known face to show poses or expressions not present in available images.
Face models can be used as generative face detectors, too, with applications such as auto-focusing a camera
or detecting intruders on a security camera.
Face modeling can also aid face recognition, by aligning the images to be recognized or by providing the lower-dimensional representation that can be matched against a dictionary.}
\par{Variations in the conditions (such as illumination) of images present challenges for face models.
One such variation is the blurriness of photographs.
Digital cameras, particularly those in many mobile phones, are used frequently to produce photos that may not be appropriately sharp.
Even professional photographers using high-grade cameras can produce images with blurred faces in the background.
The blurriness of a facial image changes one's expectations for the face's appearance, as well as the types of variation in the appearance, so we modeled facial images using a parameter based on blurriness.}
\par{To quantify blurriness, we assumed that a Gaussian blur filter could approximate the transformation from an unobserved, unblurred image to the observed, blurred image.
This Gaussian blur is a convolution using the kernel $K(x,y \vert \sigma)$ from Equation \eqref{eq:gaussiankernel}, where $x$ is the horizontal distance and $y$ is the vertical distance between the two pixels involved.
We used $\sigma$ from Equation \eqref{eq:gaussiankernel} (with $\sigma=0$ for an unblurred image) as the PPCA parameter, because higher values of $\sigma$ create blurrier images.}
\begin{equation}
\label{eq:gaussiankernel}
K(x,y \vert \sigma) = \frac{1}{2\pi\sigma^2} \exp{\left(-\frac{x^2+y^2}{2\sigma^2}\right)}
\end{equation}

\par{We treated the facial images from the CBCL Face Database \#1 \cite{mitcbcl} as unblurred, and added Gaussian blur with a $7 \times 7$ kernel and varied $\sigma$.
The database had 472 faces chosen as the test set, and the remaining 2,429 as the training set.
We used three bins for $\sigma$: 0-1, 1-2, and 2-3.
The training and test images were created from the original faces such that each original face produced one blurred image for each of the three bins.
The parameter $\sigma$ for each observation was selected randomly from a $U(0,1)$, $U(1,2)$, or $U(2,3)$ distribution, depending on which bin's image was being produced.}
\par{We used 10 basis vectors for each IPCA bin or PPCA bin endpoint, and for PCA. For SPCA we used 30 principal vectors with 50\% nonzero entries each.
For PPCA, we used gradient descent for both the mean and basis vectors.
We used $\lambda_{m}=0.6$, $\lambda_{v}=2$, $\lambda_o=1000$, $n_c= 300$, $n_m=100$, $n_v=100$, $\alpha_m=0.01$, and $\alpha_v=0.0001$.
The training set varied from 2 to 200 examples per bin.
For all sizes of training set, all bins had images made from the same faces, but had different added blur
according to the values $\sigma$.
The smaller training sets were always subsets of the larger training sets, to allow for better examination of the
effect of the training set size.}
\begin{figure}
\centering
\includegraphics[scale=0.4]{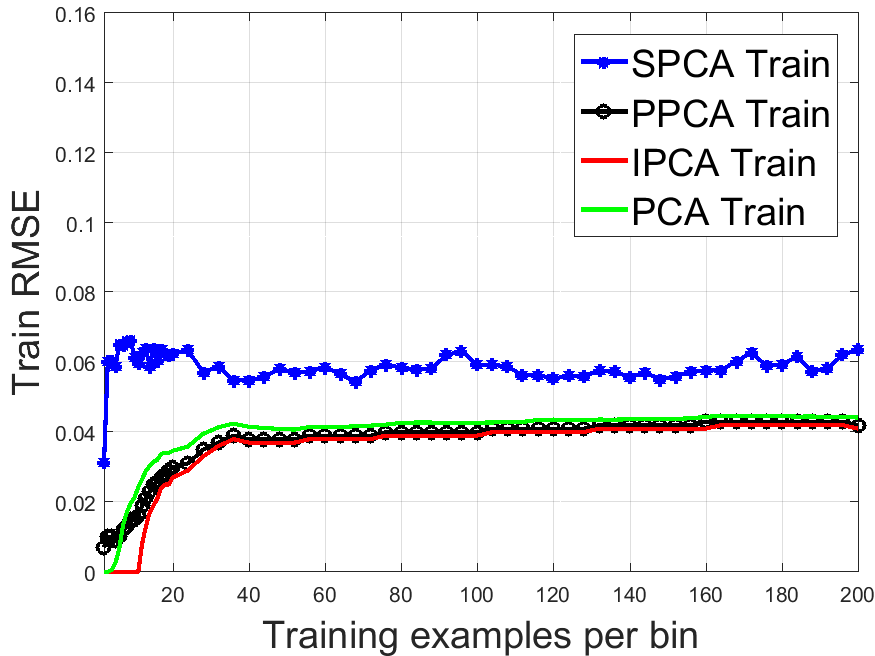}
\includegraphics[scale=0.4]{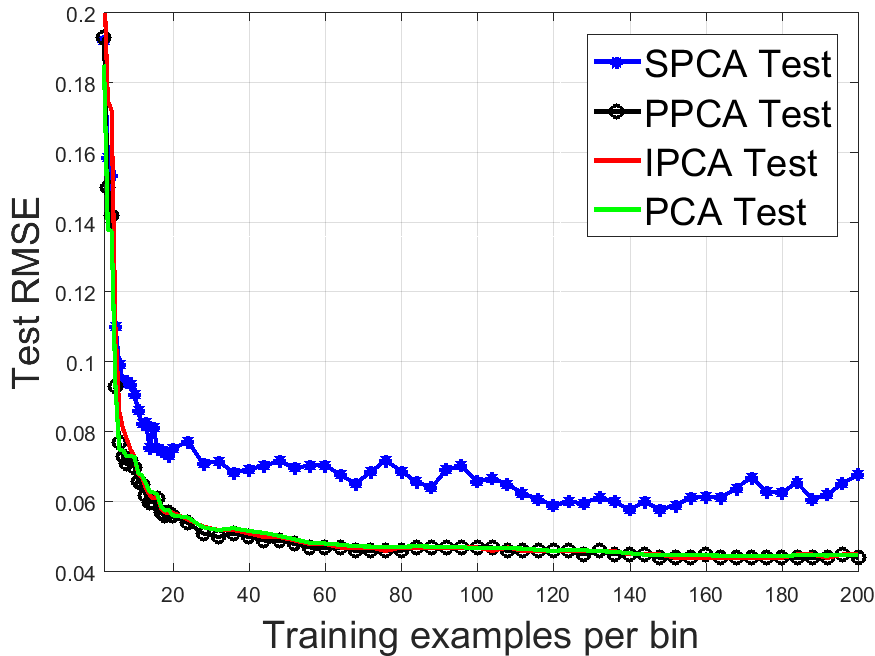}
\vskip -3mm
\caption{Mean RMSE for modeling blurred facial images, using varied numbers of training examples (Left: train set, Right: test set)}
\label{fig:rmseblur}
\vspace{-3mm}
\end{figure}

Figure \ref{fig:rmseblur} shows the mean across training or test set images for the RMSE of the blurred images' projections. Some of the errors are also summarized in Table \ref{tab:rmseblur}.
\begin{table}[htb]
\vspace{-3mm}
\small
\begin{center}
\caption{Summary of RMSE results for the blur data.}\label{tab:rmseblur}
%\vskip -1mm
\begin{tabular}{|c|cccc|cccc|}
\hline
Training examples    &\multicolumn{4}{c}{Train RMSE} &\multicolumn{4}{|c|}{Test RMSE} \\
per bin  &PCA &IPCA &PPCA &SPCA &PCA &IPCA &PPCA &SPCA\\
\hline
2   &0.000 &0.000	&0.007 &0.031 &0.185 &0.211	&0.193 &0.192\\
\hline
%8 &0.017 &0.000	&0.013 &0.111 &0.073 &0.078	&0.071  &0.138\\
10 &0.021 &0.000	&0.015 &0.062 &0.073 &0.073	&0.070  &0.091\\
\hline
%15 &0.032 &0.019	&0.025 &0.106 &0.063 &0.061	&0.060 &0.120\\
20 &0.035 &0.027	&0.030  &0.063&0.056 &0.057	&0.056 &0.075\\
\hline
%40 &0.042 &0.037	&0.038 &0.085 &0.052 &0.051	&0.050 &0.101\\
50 &0.041 &0.037	&0.038 &0.057 &0.050 &0.050	&0.049 &0.070\\
\hline
%60 &0.042 &0.038	&0.039 &0.088 &0.048 &0.048	&0.047 &0.101\\
100 &0.043 &0.039	&0.040 &0.059 &0.047 &0.047 &0.047 &0.066\\
\hline
200 &0.044 &0.041	&0.042 &0.064 &0.045 &0.045 &0.044 &0.068\\
\hline
\end{tabular}
\end{center}
\vspace{-6mm}
\end{table}

PPCA had lower approximation error on the test set than IPCA for each training size from 2 to 200 examples per bin,
but had a more noticeable advantage when both methods used eight or fewer training examples per bin.
SPCA had large training and testing errors, sign that the sparse model cannot fit well this kind of data.
PCA did a very good job, comparable to IPCA and PPCA.

\noindent{\bf Face Recognition.} Even though PPCA is designed for manifold approximation and not for classification, we would like to see how these methods compare for face recognition. We can evaluate face recognition performance on this data, since we have three versions of each face, with different blur values. 

We experimented with two data sizes, with  100 and 200 faces, each having three blurred versions of each face. For each face, from the three available versions we chose one at random for testing and the other two for training. 

We also evaluated two other manifold methods: Modified PCA (MPCA)\cite{luo2003modified}, which is an modification of PCA for recognition that normalizes the PC coefficients by dividing them to the square root of their corresponding eigenvalues, and Locality Preserving Projections \cite{niyogi2004locality}, which finds a linear projection of the data so that most of its local information is preserved.

Each method was used to learn a low dimensional representation  and the training observations were projected to this low dimensional space.
Given a test face, it was projected to the low dimensional space and the training observation of maximal correlation was used to obtain the recognition result. The dimension $d$ of the low dimensional space was chosen for each method from $d\in \{10,20,...,100\}$ to obtain the smallest average test error over 100 random splits of the training/test data.
%of dimension 10 for PCA, PPCA and IPCA, and dimension 30 for SPCA)

\begin{table}[htb]
\vspace{-5mm}
\small
\begin{center}
\caption{Recognition errors averaged over 100 random splits of the training/test sets.}\label{tab:recog}
%\vskip -1mm
\begin{tabular}{|c|c|c|c|c|c|c|}
\hline
Faces 	&PCA &MPCA\cite{luo2003modified} &SPCA\cite{zou2006sparse} &IPCA &PPCA &LPP\cite{niyogi2004locality}\\
\hline
100   &4.94 (1.50) &3.33 (1.30)  &5.25 (1.62)&72.78 (8.55) &12.33 (2.45) &62.59 (4.76) \\
200   &6.50 (1.19) &4.98 (1.35)  &6.33 (1.17) &72.69 (10.58) &14.67 (3.36) &77.07 (2.72)\\
\hline
\end{tabular}
\end{center}
\vspace{-6mm}
\end{table}

In Table \ref{tab:recog} are show the detection rates for the datasets with 100 and 200 faces. The best recognition errors are obtained by MPCA, followed by PCA and SPCA. PPCA comes next, doing a much better job than IPCA and LPP.

\subsection{Facial Images with Rotation}
\label{section:yaw}
\par{Section \ref{section:blur} addressed the challenges of modeling facial images with different levels of blurriness.
A separate challenge in face modeling is out-of-plane rotation, which changes the expected appearance of facial features and produces predictable changes in the occlusion of important facial features.
Yaw rotation is highly prevalent in photos, particularly for ``in the wild" photos, which are taken in uncontrolled settings, often by ordinary users.
One could model pitch or roll rotation with PPCA, but we focus on yaw rotation because it has the largest variation in the available face images.}
\subsubsection{Background}
\par{Linear models for facial appearances exist, such as active appearance models (AAMs) \cite{cootes2001active} and 3D morphable models (3DMMs) \cite{blanz1999morphable}.
AAMs typically incorporate in-plane rotation and suffer from an inability to model out-of-plane rotation, but 3DMMs exist in 3D and can use 3D rotation.
3DMMs can be fit to 2D test images, but they are trained using 3D facial scans performed in a laboratory setting.
Potential users typically do not have the necessary equipment for these scans, and even with the equipment, one has very limited training data relative to a dataset of 2D images.
Furthermore, applications for in-the-wild images grow as these images become more important for social media and other Internet uses,
and the laboratory setting on the scan data makes them dissimilar to in-the-wild images.
Zhu and Ramanan (2012) showed that training on in-the-wild images greatly increases face detection performance on in-the-wild test data \cite{zhu2012face}, and it seems logical that similarity between training and test data would be desirable for face modeling as well.}
\par{Gross, Matthews, and Baker (2004) modify AAMs to address occlusion \cite{gross2004constructing}.
This does not distinguish occlusion by an object (such as a hand in front of a face) from self-occlusion caused by out-of-plane rotation,
so it does not take advantage of the more predictable nature of rotation-based self-occlusion.
Xiao et al. (2004) also modify AAMs, creating a hybrid of a 2D and a 3D model by adding extra parameters and constraints to a 2D model \cite{xiao2004real}.
It allows the training advantages of a 2D model with some of the advantages of a 3D model, but compared to PPCA,
it does not address out-of-plane rotation as directly and relies more on 3D elements not directly observable in the 2D data.}
\par{AAMs and 3DMMs each incorporate two linear models: one for the shape mesh, and one for the appearance after removing the influence of shape variation.
AAMs typically use frontal images only and translate the appearance from the original image's shape mesh to the mean shape mesh by triangular warping.
Yaw rotation creates predictable changes to both the shape and the appearance.
PPCA could model both, but we chose to focus on the appearance component, and modeled the shape using a rigid, 3D shape model built on other data.}

\subsubsection{Data}
\par{We used 272 human facial images from the Annotated Facial Landmarks in the Wild (AFLW) database
\cite{koestingeraflw}, which includes annotations of the locations and occlusion status of 21 key points.
We chose the subset such that the faces were all in color and appeared to be of 272 different people.
We used yaw rotation in radians as the PPCA parameter $\theta$.
It was limited to the range from $-\pi/2$ to $\pi/2$, and we divided the range into 16 equally-sized bins.
Our subset of AFLW had 17 images in each bin, and three images per bin were selected randomly to be in the test set.
The remaining 14 images per bin were eligible for training, but we varied the training set size from 2 to 14 images per bin.
The smaller training sets were always subsets of the larger training sets.
Values of $\theta$ came from finding the roll, pitch, and yaw angles that best rotated the rigid shape model
to fit the unoccluded key points' horizontal and vertical coordinates.
Several yaw angles and key point locations were corrected manually.}

\begin{figure}[h]
\centering
\includegraphics[height=2cm]{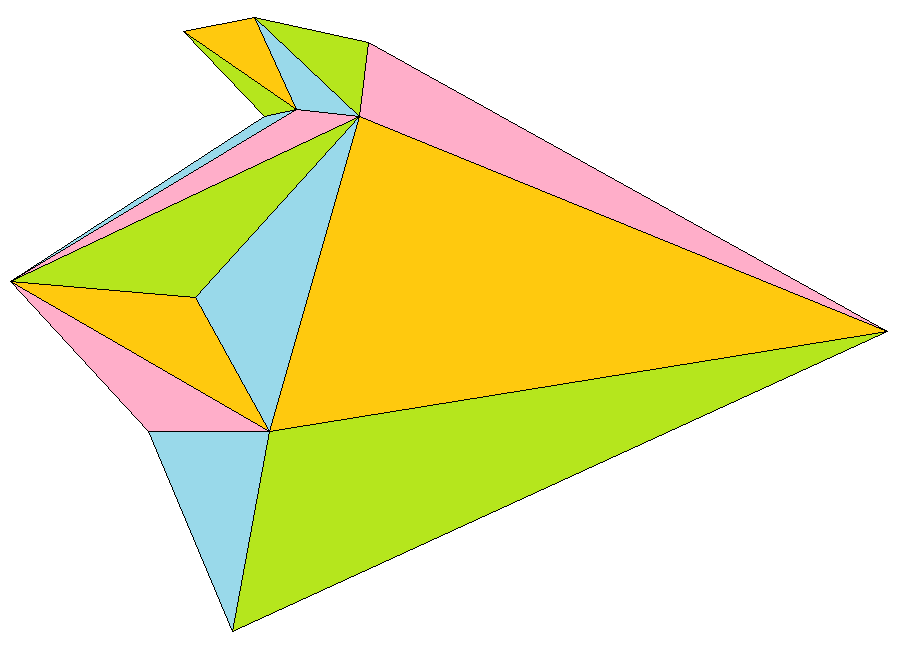}
\includegraphics[height=2cm]{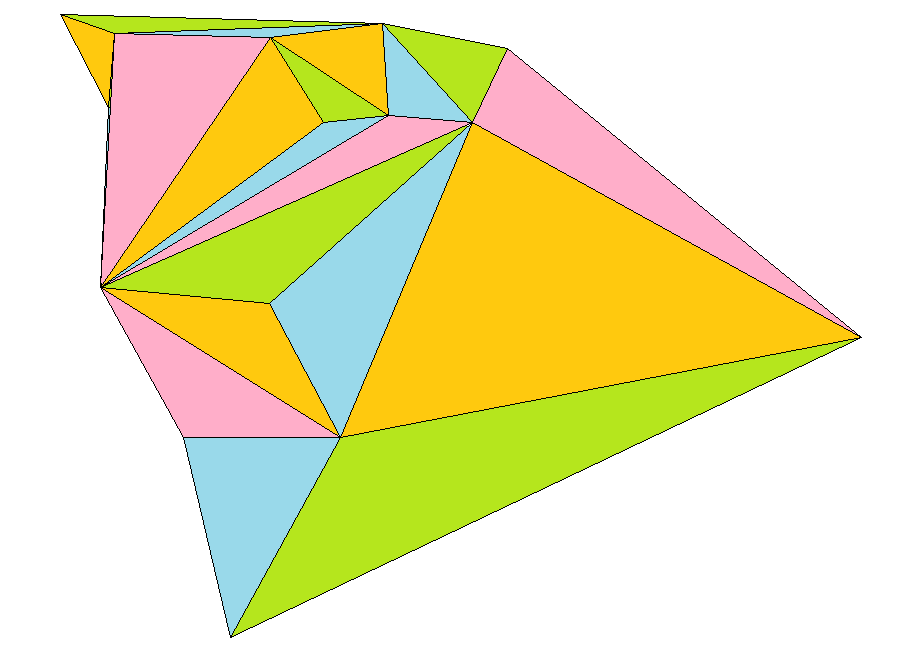}
\includegraphics[height=2cm]{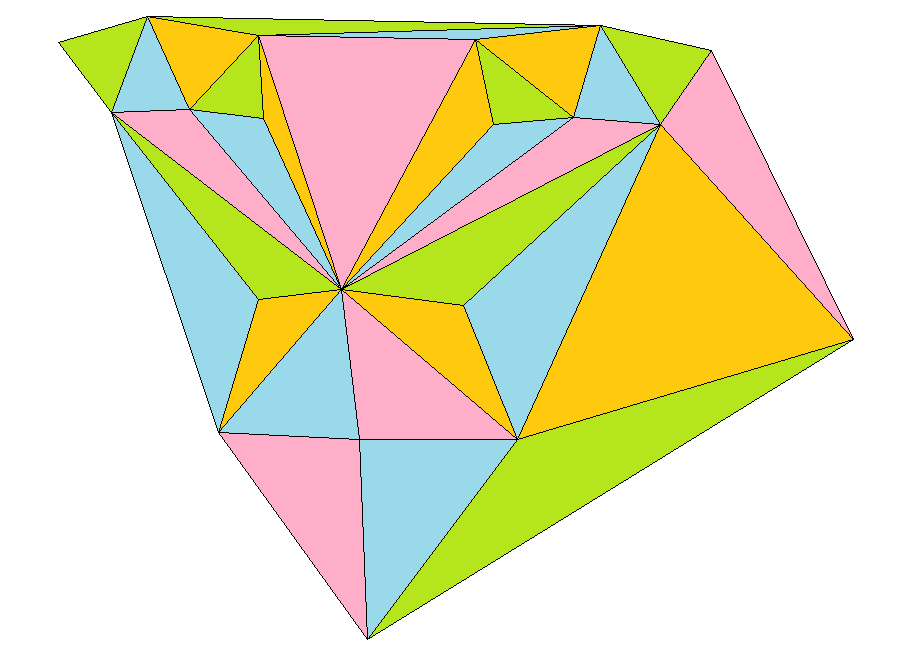}
\includegraphics[height=2cm]{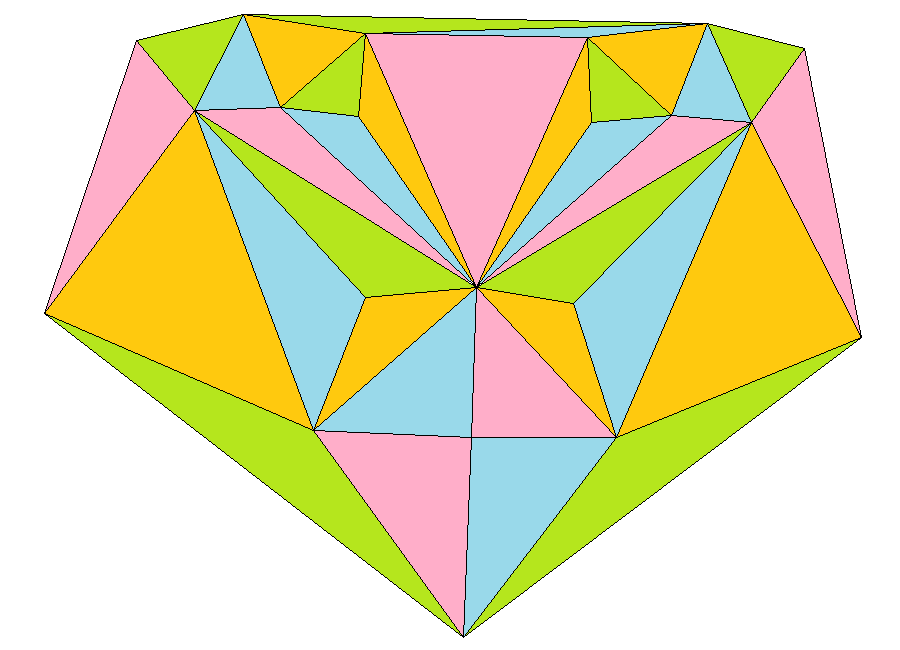}
\includegraphics[height=2cm]{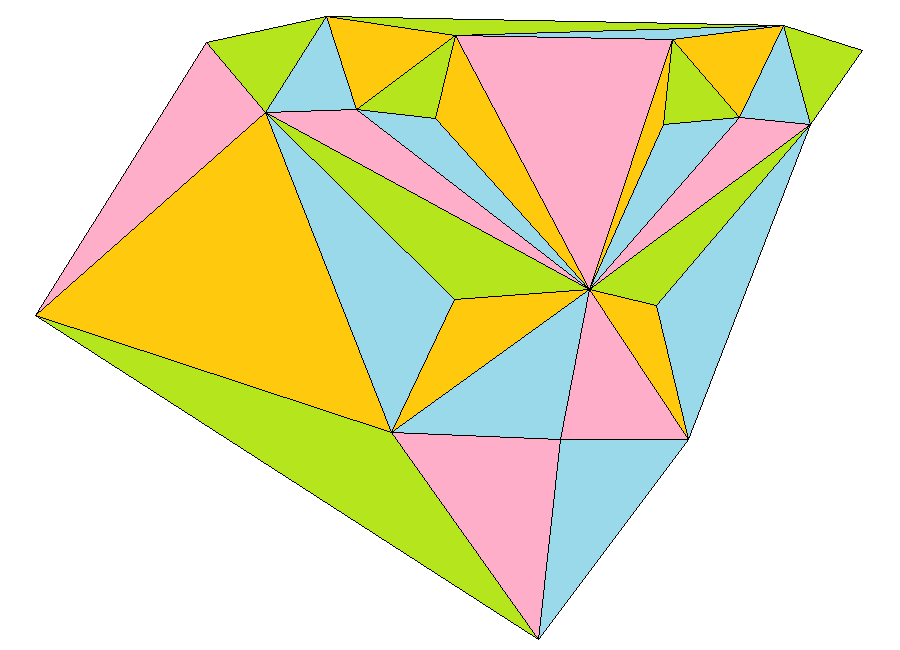}
\includegraphics[height=2cm]{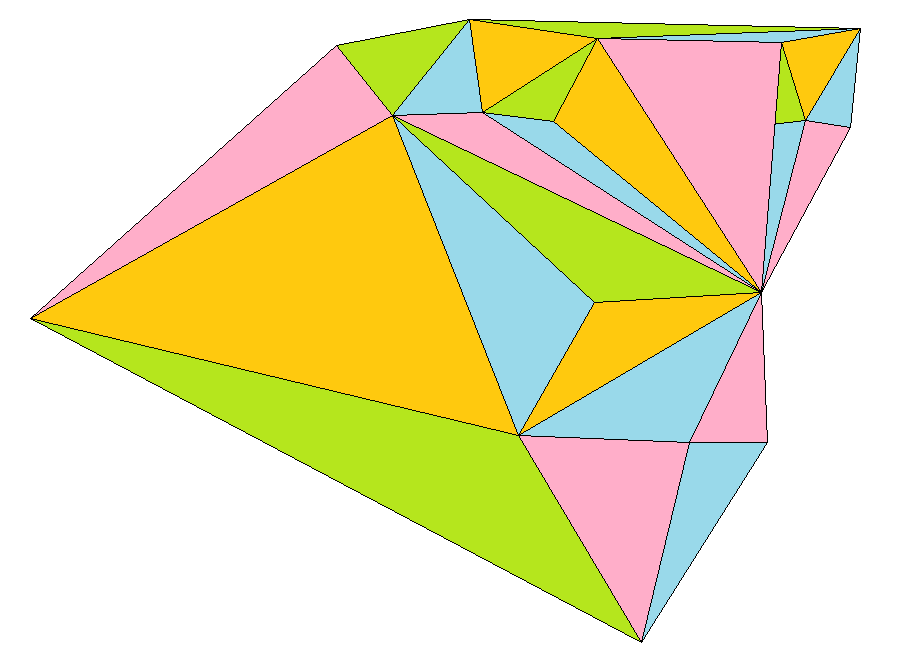}
\includegraphics[height=2cm]{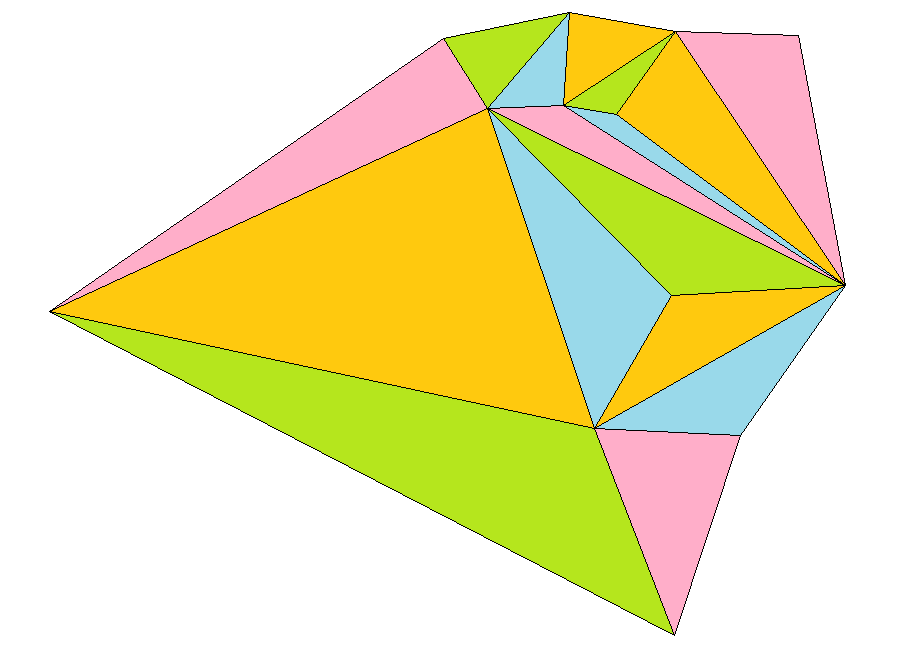}
\includegraphics[height=2cm]{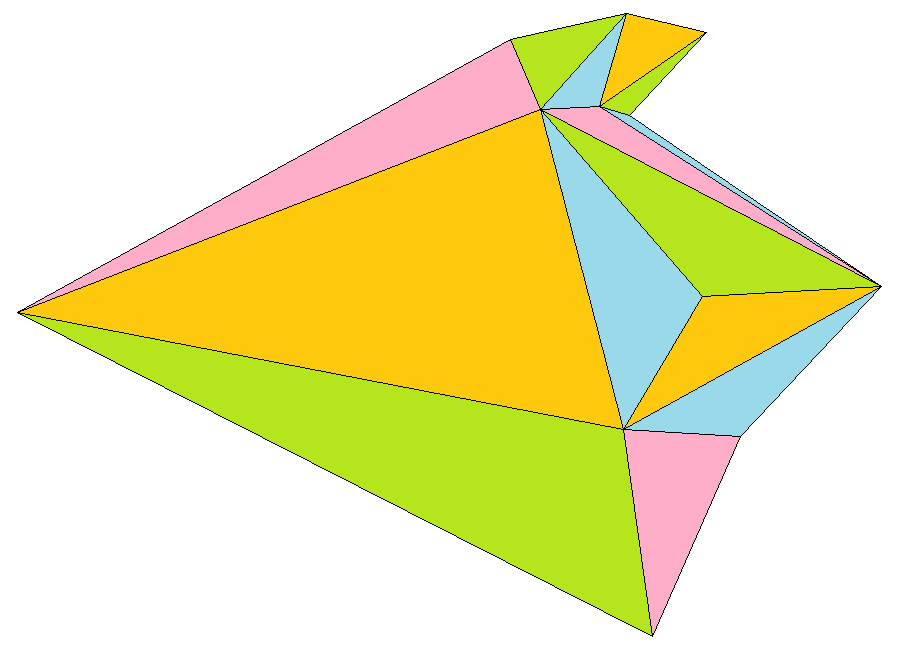}
\vskip -3mm
\caption{Triangulation at bin endpoints 2, 5, 8, 9, 10, 12, 14, and 16}
\label{fig:triangulation}
\vspace{-3mm}
\end{figure}

\par{AAMs commonly use a triangulation of the face to translate a shape mesh of key points into a shape that can cover pixels.
We also used a triangulation, which we constructed manually to have triangles that are less likely to have one of three vertices occluded at yaw angles from $-\pi/2$ to $\pi/2$.
This generally implied triangles that ran more vertically than in automatic triangulation methods.
PPCA promotes the smoothness of adjacent bin endpoints, so the triangles needed to use pixels that corresponded to equivalent areas in other bin endpoints' shapes.
We calculated the triangle's area for each bin endpoint shape in our 3D model, and used the largest-area version of the triangle for PPCA.
We warped each triangle from the original images to these model triangles, which were considered occluded or not based on the direction of the normal vector to that triangle in the rigid shape model rotated to the appropriate yaw angle.
AFLW's image-specific occlusion annotations were not used after estimating $\theta$.}

%We tested two numbers of basis vectors per bin endpoint, 4 and 10, to determine whether this modeling choice affects the appropriateness of PPCA.
%IPCA bins used the same number of basis vectors as PPCA bin endpoints did.
We used 10 basis vectors for each IPCA bin or PPCA bin endpoint, and for PCA. For SPCA we used 30 principal vectors with 50\% nonzero entries each.
We also investigated the influence of whitening.
The intensities before whitening were represented as double floating-point numbers from zero (black) to one (white).
If whitening were used, each image would get six additional parameters in its representation, which were not a part of PPCA (or IPCA) itself.
After warping an image, we stored the original mean intensity and standard deviation for red, green, and blue.
We translated and rescaled the intensities such that each color had a mean of 0.5 and a standard deviation of 0.031.
The latter was chosen to be just large enough to keep all whitened intensities within the [0, 1] interval.
PPCA and IPCA modeled the whitened versions, and after projecting the whitened image, we reversed the whitening transformation using the image-specific means and standard deviations by color.

\begin{figure}[htb]
\centering
\includegraphics[scale=0.4]{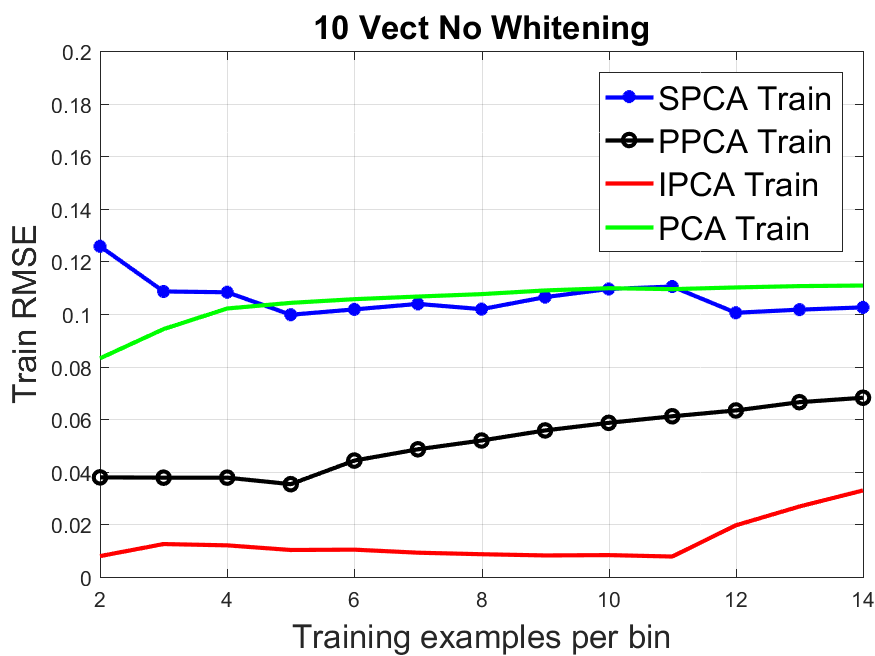}
\includegraphics[scale=0.4]{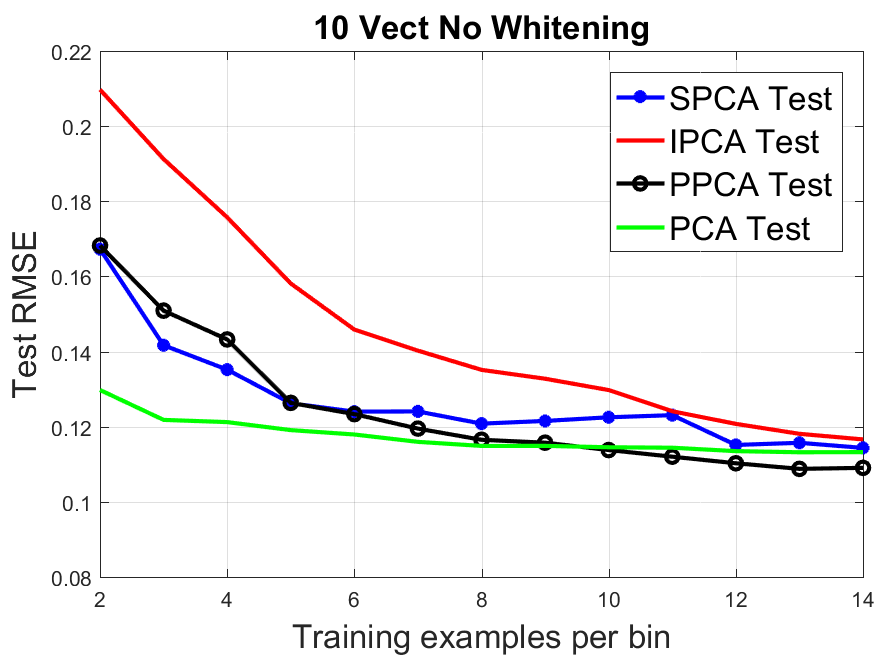}
\includegraphics[scale=0.4]{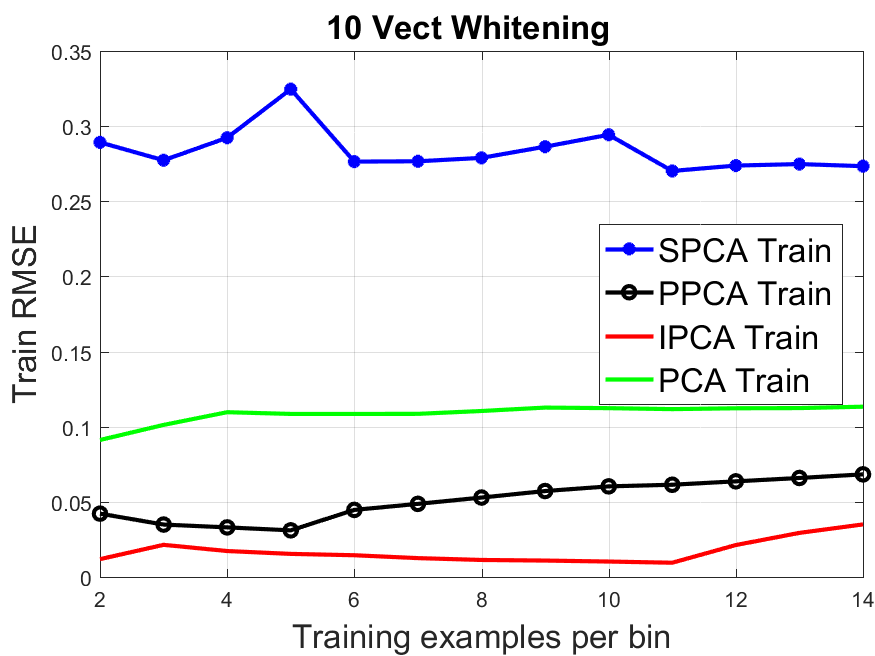}
\includegraphics[scale=0.4]{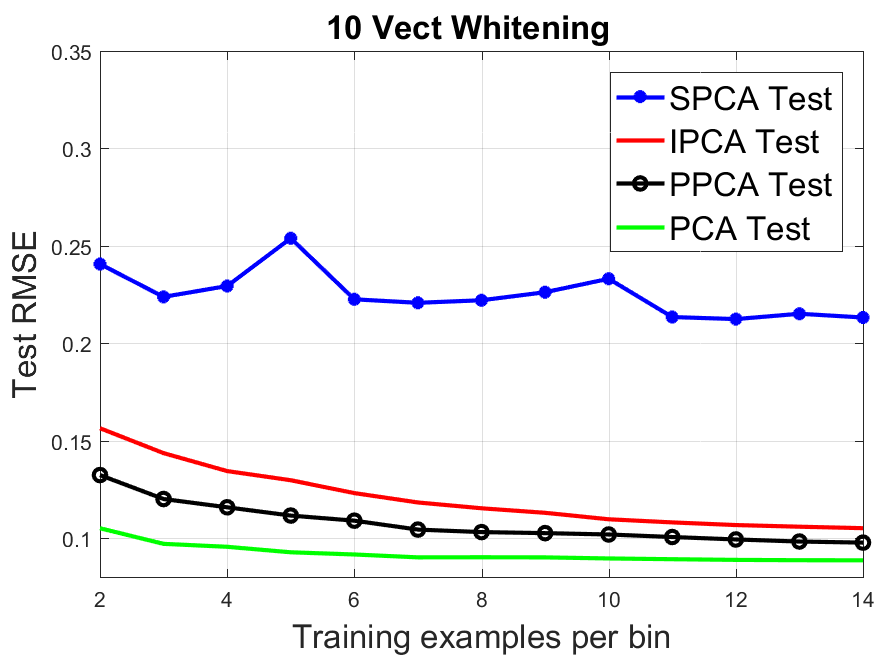}
\vskip -3mm
\caption{Mean RMSE for projection of facial images with yaw rotation parameter, evaluated on training (left) and test (right) sets using varied numbers of training examples.}
\label{fig:rmseyaw}
\vspace{-3mm}
\end{figure}

\subsubsection{Model Fitting and Results}
\par{We trained models with training set sizes from 2 to 14 examples per bin. PPCA needed to use gradient descent to optimize both the mean and basis vectors.
We used $\lambda_{m}=0.001$, $\lambda_{v}=0.01$, $\lambda_o=1000$, $n_c=200$, $n_m=100$, $n_v=250$, $\alpha_m=0.0001$, and typically $\alpha_v=10^{-6}$.
The models with two to four training examples per bin and 10 basis vectors required smaller $\alpha_v$ to avoid divergence.
The occlusion of each triangle was considered known, because it was treated as a function of a known yaw angle.
So, we set the image-specific mean vector and basis vectors in IPCA and PPCA projections to have zeros for any out-of-shape pixels before we found images' coefficients.
After projecting the image and reversing whitening if it was used, we calculated the RMSE for each image in the training and test sets.}
\par
\begin{figure}[htb]
\centering
\includegraphics[height=5cm]{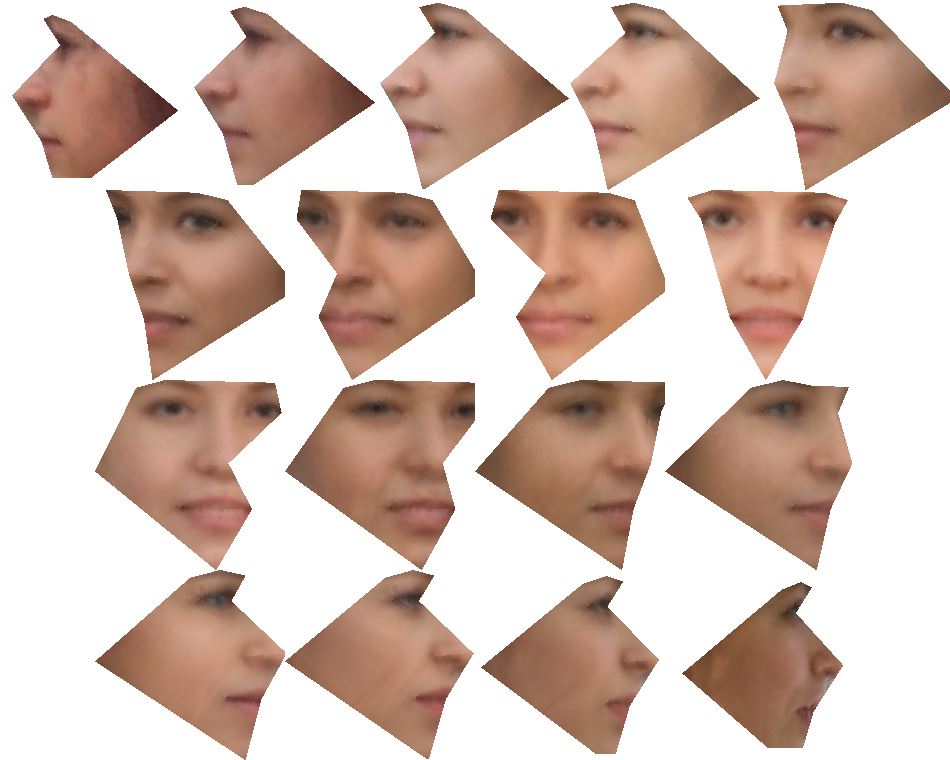}
\vskip -3mm
\caption{Mean facial images by rotation-based bin endpoint for PPCA, using higher smoothness penalty $\lambda_{m}=0.1$ and no whitening.}
\label{fig:meansyawppcahigherpenalty}
\vspace{-3mm}
\end{figure}

{Figure \ref{fig:rmseyaw} shows the means of these RMSEs, which are averaged across the images of the training or test set. Some of these results are also summarized in Table \ref{tab:rmseface}.
\begin{table}[htb]
\vspace{-1mm}
\small
\begin{center}
\caption{Summary of RMSE resuts for different methods.}\label{tab:rmseface}
%\vskip -1mm
\begin{tabular}{|l|c|c|c|c|c|c|c|c|c|}
\hline
 &  &\multicolumn{4}{c}{Train RMSE} &\multicolumn{4}{|c|}{Test RMSE} \\
%\hline
%&  &\multicolumn{2}{c}{4 Vectors} &\multicolumn{2}{|c|}{10 Vectors}&\multicolumn{2}{c}{4 Vectors} &\multicolumn{2}{|c|}{10 Vectors} \\
\hline
Data &Whitening  &PCA &IPCA &PPCA &SPCA  &PCA &IPCA &PPCA &SPCA \\
\hline
2 per bin &no &0.0831 &0.0079	&0.0378 &0.1258 &0.1298 &0.2097	&0.1682 &0.1673\\
8 per bin &no  &0.1076  &0.0086	&0.0519 &0.1018 &0.1149  &0.1351	&0.1165 &0.1208\\
14 per bin &no  &0.1108  &0.0329	&0.0681 &0.1025 &0.1132  &0.1166	&0.1090 &0.1143\\
\hline
2 per bin &yes &0.0911  &0.0118	&0.0421 &0.2892 &0.1050  &0.1563	&0.1323 &0.2407\\
8 per bin &yes &0.1104  &0.0113	&0.0528 &0.2789  &0.0901  &0.1152	&0.1030  &0.2220\\
14 per bin &yes &0.1133  &0.0349	&0.0682 &0.2734 &0.0885 &0.1050	&0.0976  &0.2132\\
\hline
\end{tabular}
\end{center}
\vspace{-6mm}
\end{table}
%2 per bin &no &0.0079	&0.0380  &0.0079	&0.0378 &0.2098	&0.1688 &0.2097	&0.1682\\
%8 per bin &no  &0.0607	&0.0922  &0.0086	&0.0519 &0.1408	&0.1282 &0.1351	&0.1165\\
%14 per bin &no  &0.0843	&0.1019 &0.0329	&0.0681 &0.1275	&0.1233 &0.1166	&0.1090\\
%\hline
%2 per bin &yes &0.0118	&0.0421 &0.0118	&0.0421 &0.1564	&0.1323 &0.1563	&0.1323\\
%8 per bin &yes &0.0581	&0.0855 &0.0113	&0.0528 &0.1183	&0.1094  &0.1152	&0.1030  \\
%14 per bin &yes &0.0778 &0.0941 &0.0349	&0.0682 &0.1117	&0.1071&0.1050	&0.0976  \\

We see that IPCA consistently overfits the data relative to PPCA.
PPCA has higher error on the training set but lower error on the test set than IPCA does.
SPCA has a hard time fitting the whitened data  but it does a better job than IPCA and slightly worse than PPCA on the data with no whitening.
PCA does a very good job on the whitened data but is outperformed by PPCA on the original data for 11-14 examples per bin.

\begin{figure}[ht]
\centering
\includegraphics[height=5cm]{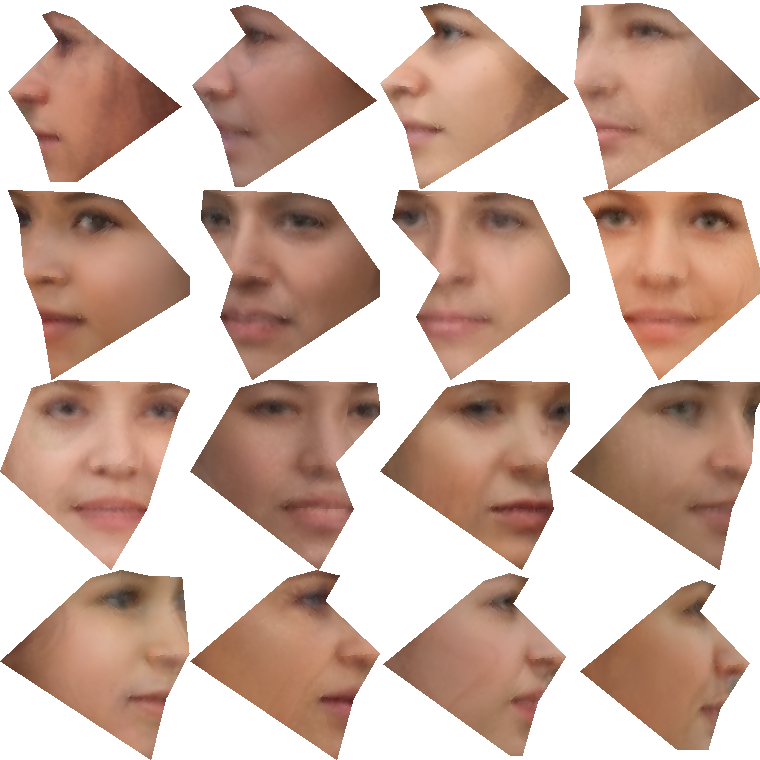}
\hspace{2mm}
\includegraphics[height=5cm]{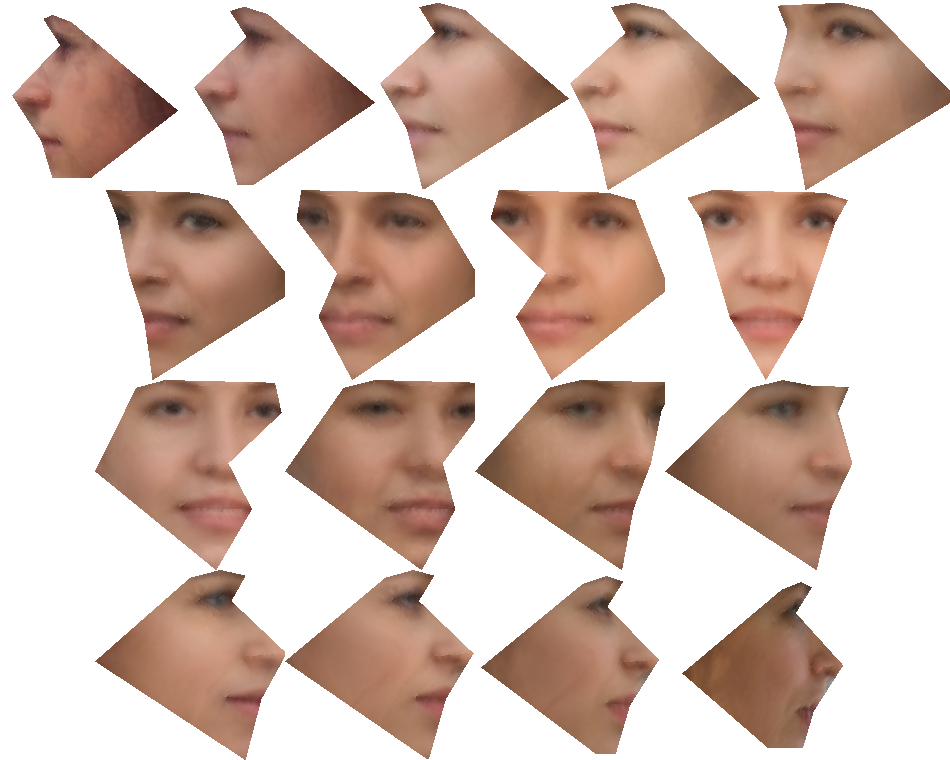}
\vskip -3mm
\caption{Mean facial images by rotation-based bin (or bin endpoint) for IPCA (left) and PPCA (right), using no whitening}
\label{fig:meansyawipcappca}
\vspace{-3mm}
\end{figure}

Figure \ref{fig:meansyawipcappca} shows the IPCA and PPCA mean vectors, warped to the bin midpoint (IPCA) or bin endpoint (PPCA) shapes.
These models used four vectors per bin (or bin endpoint), no whitening, and 12 training examples per bin.
The IPCA means appear to treat characteristics of the training images as characteristics of the bin to a higher degree than the PPCA means do.
One can see more noticeable changes from bin to bin for IPCA with respect to eye color and shape, lip color, illumination, and skin complexion.
The smoothness of the mean shape can be improved further for PPCA by increasing the penalty $\lambda_{m}$ to 0.1, as shown in Figure \ref{fig:meansyawppcahigherpenalty}.
We did not test additional training set sizes with $\lambda_{m}=0.1$, but for this example, the mean RMSE for the test set (0.1220) was effectively the same as for $\lambda_{m}=0.001$ (mean RMSE = 0.1220).
Both had lower mean RMSEs for projection error than IPCA (0.1313) did.}

\section{Conclusion and Future Direction}
We have presented a novel method, parameterized principal component analysis (PPCA), for modeling multidimensional data on linear manifolds that vary smoothly according to a contextually important parameter $\theta$.
We compared PPCA to independent principal component analysis (IPCA), which uses separate PCA models for groups formed by values of the parameter $\theta$.
We showed that PPCA outperformed IPCA at recovering known true mean vectors and true basis vectors based on smooth functions of the parameter $\theta$, at producing lower approximation error on three datasets and at obtaining smaller face recognition errors on one dataset.
These datasets contained lymph node shapes that varied by the diameter, blurred human facial images that varied by the standard deviation $\sigma$ of the Gaussian blur applied, and human facial images that varied by the angle of yaw rotation.
%In each of the three datasets, PPCA's performance on the test set was the strongest relative to IPCA when the two methods used smaller training sets.
\par{We have explored three types of applications of PPCA to datasets, with different types of parameter in each.
However, many other applications exist and future work could extend PPCA to more parameters than the three we tested.
Also, we performed some investigation of different modeling choices when modeling faces with different yaw rotation, but it would be beneficial to have further tests of how different numbers of basis vectors used and different adjustments to the data affect the utility of PPCA.}

\section*{References}
\bibliographystyle{plain}
\bibliography{references}
\end{document}